\documentclass[10pt,twocolumn,letterpaper]{article}

\PassOptionsToPackage{nameinlink}{cleveref}

\usepackage{cvpr}              
\usepackage{amssymb} %

\usepackage{microtype}

\renewcommand{\paragraph}[1]{\vspace{.5em}\noindent\textbf{#1}~~}

\usepackage{multirow} 
\usepackage{booktabs}
\usepackage{arydshln}   
\usepackage{siunitx}    
\usepackage{tikz,pgfplots}

\definecolor{cvprblue}{rgb}{0.21,0.49,0.74}
\usepackage[pagebackref,breaklinks,colorlinks,allcolors=cvprblue]{hyperref}
\usepackage[most]{tcolorbox}
\usepackage{float} 
\usepackage{breqn}
\usepackage{enumitem}
\usepackage{cuted}
\usepackage{titletoc}

\def\paperID{} 
\def\confName{CVPR}
\def\confYear{2026}

\title{Latent-Compressed Variational Autoencoder for Video Diffusion Models}

\author{
Jiarui Guan\textsuperscript{1} \quad
Wenshuai Zhao\textsuperscript{1,2}\thanks{Corresponding author} \quad
Zhengtao Zou\textsuperscript{1} \quad
Juho Kannala\textsuperscript{1,3} \quad
Arno Solin\textsuperscript{1,2} \\
\textsuperscript{1}Aalto University \quad
\textsuperscript{2}ELLIS Institute Finland \quad
\textsuperscript{3}University of Oulu \\
{\tt\small \{jiarui.guan, wenshuai.zhao, zhengtao.zou, juho.kannala, arno.solin\}@aalto.fi}
}

\begin{document}
\maketitle
\vspace*{-6em}

\begin{strip}
\vspace*{-2em}
\centering
\includegraphics[width=0.85\textwidth]{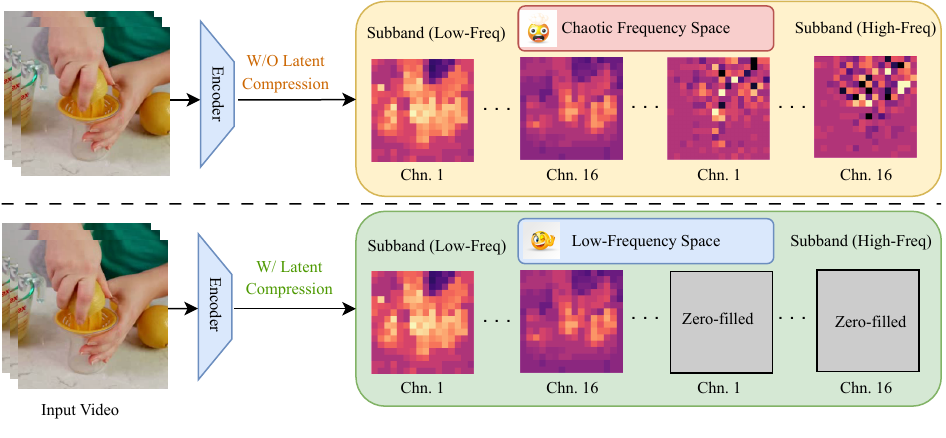}
\vspace{3mm}
\captionof{figure}{\textbf{Comparison between the schemes of video VAEs with and without the proposed latent compression.}
Our method performs frequency-aware latent compression for video generation. 
An input video is encoded and decomposed by multi-level 3D wavelet transforms (Multi-WT); low-frequency channels are retained as compact latent representations where diffusion operates. 
After denoising, the latent is zero-padded, processed by multi-level inverse wavelet transforms (Multi-IWT), and decoded into the final video. 
This design preserves global structure while reducing latent dimensionality.
}
\label{fig:teaser}
\end{strip}


\begin{abstract}
Video variational autoencoders (VAEs) used in latent diffusion models typically require a sufficiently large number of latent channels to ensure high-quality video reconstruction. However, recent studies have revealed that an excessive number of latent channels can impede the convergence of latent diffusion models and deteriorate their generative performance, even when reconstruction quality remains high. We propose a latent compression method that removes high-frequency components in video latent representations rather than directly reducing the number of channels, which often compromises reconstruction fidelity. Experimental results demonstrate that the proposed method achieves superior video reconstruction quality compared to strong baselines while maintaining the same overall compression ratio. 
\end{abstract}

\section{Introduction}
\label{sec:intro}
The success of latent diffusion models (LDMs, \cite{rombach2022high}) has established a powerful paradigm for efficient high-resolution image synthesis, where an autoencoder compresses images into a compact latent space~\cite{kingma2013auto}, allowing diffusion models to operate with reduced computational cost~\cite{hoogeboom2023simple}. This latent representation not only provides an efficient compression mechanism but also defines the perceptual upper bound of the entire generative process~\cite{rombach2022high}. Building on this foundation, recent progress has extended the latent-diffusion paradigm from static imagery to dynamic video generation. With the emergence of large-scale video generators such as Sora \cite{brooks2024video} and latent video diffusion models (LVDMs) including Open-Sora \cite{zheng2024open}, CogVideoX \cite{yang2024cogvideox}, and MovieGen \cite{polyak2024movie}, latent representation learning for videos has become a central research focus.

However, designing a latent space for videos introduces unique challenges different from those in image modeling. Prior studies have shown that maintaining a large number of latent channels is critical for achieving high reconstruction quality in variational autoencoders (VAEs)~\cite{chen2024deep}. Yet increasing the channel dimensionality substantially raises the computational cost for diffusion models and enlarges the latent search space, thereby complicating training~\cite{chen2025dc}. Furthermore, unlike images, videos require the latent space to capture both spatial fidelity and temporal coherence—an inherently difficult balance for conventional VAEs to achieve~\cite{wu2025improved}. To mitigate temporal redundancy, several recent approaches~\cite{lai2021video, yu2024efficient, liu2025hi} decompose the latent space into content and motion components or introduce specialized causal convolution operations~\cite{chen2025dc_vid, agarwal2025cosmos}. In contrast, we adopt an orthogonal perspective by closely analyzing the latent representations learned by VAE models themselves. \looseness-1

\begin{figure}[t]
  \centering
  \begin{subfigure}[t]{1\linewidth}
    \centering
    \includegraphics[width=0.96\linewidth]{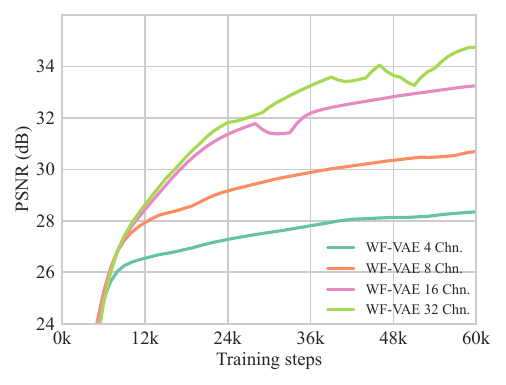}
    
  \end{subfigure}
  \caption{\textbf{Validation PSNR curves of WF-VAE~\cite{li2025wf} for 4, 8, 16, and 32 latent channels.} Increasing the number of channels yields only marginal PSNR gains, indicating substantial redundancy in the latent representation.}
  \label{fig:multi_channel_comparison}
  \vspace{-2mm}
\end{figure}

Recent studies have shown that the latent representations learned by existing autoencoders contain inordinately large high-frequency components~\cite{skorokhodov2025improving, chen2025dc}, particularly in models with a high number of latent channels. These high-frequency components can impair diffusion performance, as they conflict with the coarse-to-fine nature of the diffusion synthesis process~\cite{skorokhodov2025improving}. We further observe that the reconstruction quality gain diminishes as the number of latent channels increases, as illustrated in~\cref{fig:multi_channel_comparison}, consistent with findings from prior works~\cite{li2025wf, chen2025dc}. This behavior suggests that the latent representations contain redundant information that can be further compressed.


Inspired by the above insights, we propose a latent-compressed variational autoencoder (LC-VAE) that compresses the video latent by removing its uninformative high-frequency components, as shown in~\cref{fig:teaser}. LC-VAE performs multiple levels of wavelet decomposition in the latent space and then zeros out the high-frequency subbands of the video latent while retaining only the low-frequency subbands. This approach allows the model to store only the non-zero subbands, thereby achieving efficient latent compression while preserving the essential information required for high-quality video reconstruction. Moreover, the compact and low-frequency video representation learned by LC-VAE can further benefit downstream diffusion training. 

Extensive experimental results demonstrate that LC-VAE consistently attains higher reconstruction quality than strong baselines across multiple datasets under equivalent compression ratios. Furthermore, latent diffusion models trained on LC-VAE representations exhibit superior generation performance compared to baseline methods. The project page is available at \url{https://1mather.github.io/LC-VAE/}.

\section{Related Work}
\label{sec:related}
We first review the widely used video autoencoders for latent diffusion models. Next, we discuss prior works that incorporate wavelet transformations within video VAEs. Lastly, we introduce latent video diffusion models that are trained in video latent spaces.

\paragraph{Variational Autoencoders for Video Generation.} As autoencoders define the latent space of LDMs, their design directly influences both quality and efficiency~\cite{rombach2022high}, establishing it as a central research focus in video generation. Early latent video diffusion models typically reuse image autoencoders to compress each frame independently~\cite{blattmann2023stable, guo2023animatediff}, overlooking the temporal redundancy inherent in video data. More recent approaches incorporate spatiotemporal compression, achieving higher compression ratios by leveraging temporal coherence across frames~\cite{zhou2024allegro, yang2024cogvideox, zheng2024open, kong2024hunyuanvideo, hacohen2024ltx}. In particular, causal 3D convolutions have been integrated into autoencoders, enabling joint training on both image and video datasets~\cite{yu2023language}. Beyond the design of downsampling operators~\cite{yu2023language}, recent works have also explored alternative architectures~\cite{vaswani2017attention} and training objectives to enhance video representation learning~\cite{wu2025h3ae, agarwal2025cosmos}. While convolutional backbones remain the dominant choice for video autoencoders~\cite{blattmann2023stable}, transformer-based architectures have emerged as a promising alternative for scalable latent representation learning~\cite{hansen2025learnings}. Conventional training objectives often combine L1 reconstruction loss, Kullback–Leibler (KL) regularization, and auxiliary losses such as LPIPS and adversarial losses~\cite{rombach2022high}. However, new objectives have been proposed to further enhance reconstruction fidelity and compression efficiency~\cite{wu2025h3ae, agarwal2025cosmos}. Our approach is agnostic to both architecture and objective design, and is therefore compatible with existing advancements in the field.

\paragraph{Wavelet Transform in Video Autoencoders.} Wavelet transform~\cite{graps1995introduction} has been widely recognized as an effective technique for decomposing images and videos into different frequency subbands, and has thus been extensively utilized for compression~\cite{skodras2002jpeg, zhang2023survey}. More recently, several studies have incorporated wavelet transforms into video VAEs by first converting raw video data into the wavelet domain, thereby reducing both spatial and temporal resolutions to facilitate more efficient autoencoder processing~\cite{lin2024open, agarwal2025cosmos}. WF-VAE~\cite{li2025wf} further integrates the wavelet energy flow into deeper layers of the VAE, enriching the learned representations and achieving faster convergence with superior performance. In contrast to existing approaches that apply wavelet transforms directly to video pixel spaces, we perform the transform on latent representations. This enables the removal of high-frequency latent components, thereby enhancing compression efficiency and improving the stability of diffusion model training~\cite{skorokhodov2025improving}.

\paragraph{Latent Video Diffusion Models.} While autoregressive video generation models~\cite{yu2023language, kondratyuk2023videopoet} synthesize videos within discrete token spaces, latent diffusion models~\cite{he2022latent, blattmann2023stable} operate in continuous latent spaces and form the backbone of several state-of-the-art text-to-video models~\cite{huggingface_video_generation_leaderboard}, including Sora~\cite{brooks2024video}, Hunyuan Video~\cite{kong2024hunyuanvideo}, Wan~\cite{wan2025wan}, and Seedance 1.0~\cite{gao2025seedance}. Beyond the design of the video VAEs~\cite{yu2023language}, these models often introduce specific diffusion model architectures~\cite{kong2024hunyuanvideo}, employ distinct training strategies~\cite{hacohen2024ltx} or perform specific data curation~\cite{zhou2024allegro}. In this work, however, we focus on developing a diffusion-agnostic video VAE, rather than pursuing state-of-the-art text-to-video generation, which typically requires models with billions of parameters. Accordingly, we evaluate our proposed LC-VAE using both lightweight latent diffusion models~\cite{ma2025latte} and a large-scale model~\cite{wan2025wan}.

\section{Background: 3D Wavelet Transform}

The 3D wavelet transform provides an effective framework for decomposing video data into multi-scale frequency components across spatial and temporal dimensions. For a video sequence $\mathbf{v} \in \mathbb{R}^{C \times T \times H \times W}$ with $C$ channels, $T$ temporal frames, and spatial dimensions $H \times W$, the 3D Haar wavelet decomposition~\cite{talukder2010haar} employs two fundamental one-dimensional filters: a low-pass averaging filter $\phi = \frac{1}{\sqrt{2}}[1, 1]$ and a high-pass detail-capturing filter $\psi = \frac{1}{\sqrt{2}}[1, -1]$. These orthogonal filters operate along each dimension to extract complementary information, where the low-pass filter captures smooth variations and approximations while the high-pass filter detects local discontinuities and fine-scale details.

At decomposition level $\ell$, the 3D wavelet transform $\mathcal{W}$ generates eight distinct frequency subbands by applying combinations of $\phi$ and $\psi$ along the temporal, height, and width axes. Formally, each subband $\mathbf{B}^{(\ell)}_{abc} \in \mathbb{R}^{C \times T_\ell \times H_\ell \times W_\ell}$ at level $\ell$ is computed by decomposing its upper level subbands:
\begin{equation}
\mathbf{B}^{(\ell)}_{abc} = \mathcal{W}(\mathbf{B}^{(\ell-1)})= \mathbf{B}^{(\ell-1)} \circledast (\xi_a \otimes \xi_b \otimes \xi_c)
\label{equ:wavelet}
\end{equation}
where $\xi_a, \xi_b, \xi_c \in \{\phi, \psi\}$ denote the filters applied along the temporal, height, and width dimensions respectively, $\circledast$ represents the convolution operation, and $\otimes$ means tensor product. $\mathbf{B}^{(0)} = \mathbf{v} \in \mathbb{R}^{C \times T \times H \times W}$ represents the initial video to decompose. The downsampling inherent in the wavelet transform reduces dimensions such that $T_\ell = T_{\ell-1}/2$, $H_\ell = H_{\ell-1}/2$, and $W_\ell = W_{\ell-1}/2$. The notation $abc$ uses $\textrm{L}$ (low-pass) and $\textrm{H}$ (high-pass) to indicate the filter type for each dimension. 

Consequently, the complete set of subbands at level $\ell$ is 
\begin{align}
\mathcal{B}^{(\ell)} = \{&\mathbf{B}^{(\ell)}_{\textrm{LLL}}, \mathbf{B}^{(\ell)}_{\textrm{LLH}}, \mathbf{B}^{(\ell)}_{\textrm{LHL}}, \mathbf{B}^{(\ell)}_{\textrm{\textrm{HLL}}}, \notag\\
&\mathbf{B}^{(\ell)}_{\textrm{LHH}}, \mathbf{B}^{(\ell)}_{HHL}, \mathbf{B}^{(\ell)}_{\textrm{HLH}}, \mathbf{B}^{(\ell)}_{\textrm{HHH}}\},
\label{equ:basic_wavelet_subbands}
\end{align}
where each subband has shape $\mathbb{R}^{C \times T_\ell \times H_\ell \times W_\ell}$. Multi-level decomposition proceeds hierarchically by recursively applying the transform to upper-level subbands: $\mathbf{B}^{(\ell)}_{abc} \rightarrow \mathcal{B}^{(\ell+1)}$, yielding a pyramidal representation where the decomposed subband at level $\ell$ has dimensions reduced by a factor of $2^\ell$ in each axis, \ie, $\mathbf{B}^{(\ell)}_{abc} \in \mathbb{R}^{C \times T/2^\ell \times H/2^\ell \times W/2^\ell}$.

\section{Method}


In~\cref{sec:analysis}, we analyze the frequency distribution of the latent representation in existing video autoencoders, which motivates our work to filter out uninformative frequency components in the latent representations. Then, in~\cref{sec:architecture} and \cref{sec:training}, we describe the detailed architecture and training of the proposed latent-compressed video autoencoder.

\subsection{Analysis of Latent Representations}
\label{sec:analysis}

\begin{figure}[t]
  \centering
  \includegraphics[width=\linewidth, trim={0 0cm 0 0cm}, clip]{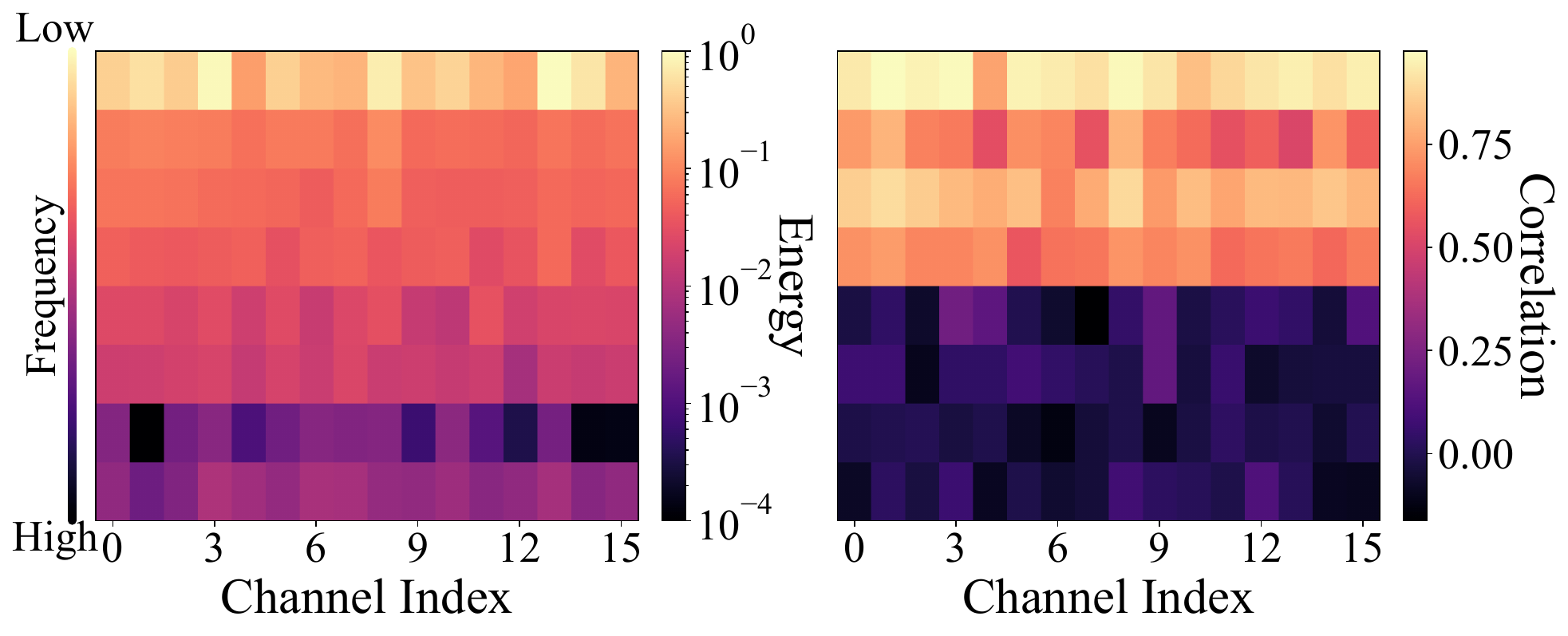}
  \caption{\textbf{Energy and correlation distribution across frequencies.} We visualize heatmaps of the normalized energy (left) and per-channel lag-1 temporal autocorrelation~\cite{sun2025adaptingllmstimeseries} (right) obtained by applying a 3D Haar wavelet transform to video latent representations encoded by WF-VAE~\cite{li2025wf}. Columns correspond to latent channels, and rows represent different frequency subbands. The visualization reveals that low-frequency subbands exhibit higher energy and stronger temporal correlation, whereas high-frequency subbands tend to be less informative and more disordered.}
  \label{fig:frequency_analysis}
\end{figure}

\paragraph{Frequency Domain Visualization.} We begin by analyzing the frequency characteristics of video representations learned by a baseline model, the Wavelet-Flow VAE (WF-VAE)~\cite{li2025wf}. Specifically, we apply a 3D wavelet transform to the video latent representation and visualize both the inter-frame correlation and energy distribution, as shown in~\cref{fig:frequency_analysis}. Formally, given a video $\mathbf{v} \in \mathbb{R}^{C \times T \times H \times W}$, its latent representation is obtained as $\mathbf{z} = \mathcal{E}_{\theta}(\mathbf{v}) \in \mathbb{R}^{C^{\prime} \times \frac{T}{t} \times \frac{H}{f} \times \frac{W}{f}}$, where $\mathcal{E}_{\theta}$ denotes the WF-VAE encoder parameterized by $\theta$. The encoder reduces the spatial resolution by a factor of $f$ and the temporal resolution by a factor of $t$, while $C^{\prime}$ represents the number of latent channels. We then perform a one-level 3D wavelet transform, as defined in~\cref{equ:wavelet}, on the latent representation $\mathbf{z}$ rather than the raw video $\mathbf{v}$, yielding eight subbands that correspond to distinct frequency components of the video latent.

To quantify the temporal consistency of each frequency subband, we compute the lag-1 temporal autocorrelation~\cite{sun2025adaptingllmstimeseries} for each subband and visualize the results in~\cref{fig:frequency_analysis} (right). The lag-1 autocorrelation measures the linear dependency between consecutive frames within a subband, reflecting how smoothly the signal evolves over time. The low-frequency subbands, such as $\mathbf{B}_{\textrm{LLL}}$, $\mathbf{B}_{\textrm{LLH}}$, and $\mathbf{B}_{\textrm{LHL}}$, exhibit significantly higher temporal autocorrelation compared to higher-frequency subbands. In~\cref{fig:frequency_analysis} (left), we further present the energy distribution across subbands, where low-frequency components dominate the overall energy. These observations suggest that low-frequency components primarily capture the structural information of the video content which is more temporally correlated, while high-frequency components encode finer details such as textures and noise~\cite{skorokhodov2025improving}. Based on this insight, we hypothesize that structural information can be efficiently compressed into the video latent representation, while offloading the recovery of detailed textures to the decoder conditioned on the encoded low-frequency content.

\begin{figure*}[ht] \centering\footnotesize 
\includegraphics[width=0.98\textwidth, trim={0 0cm 0 0cm}, clip]{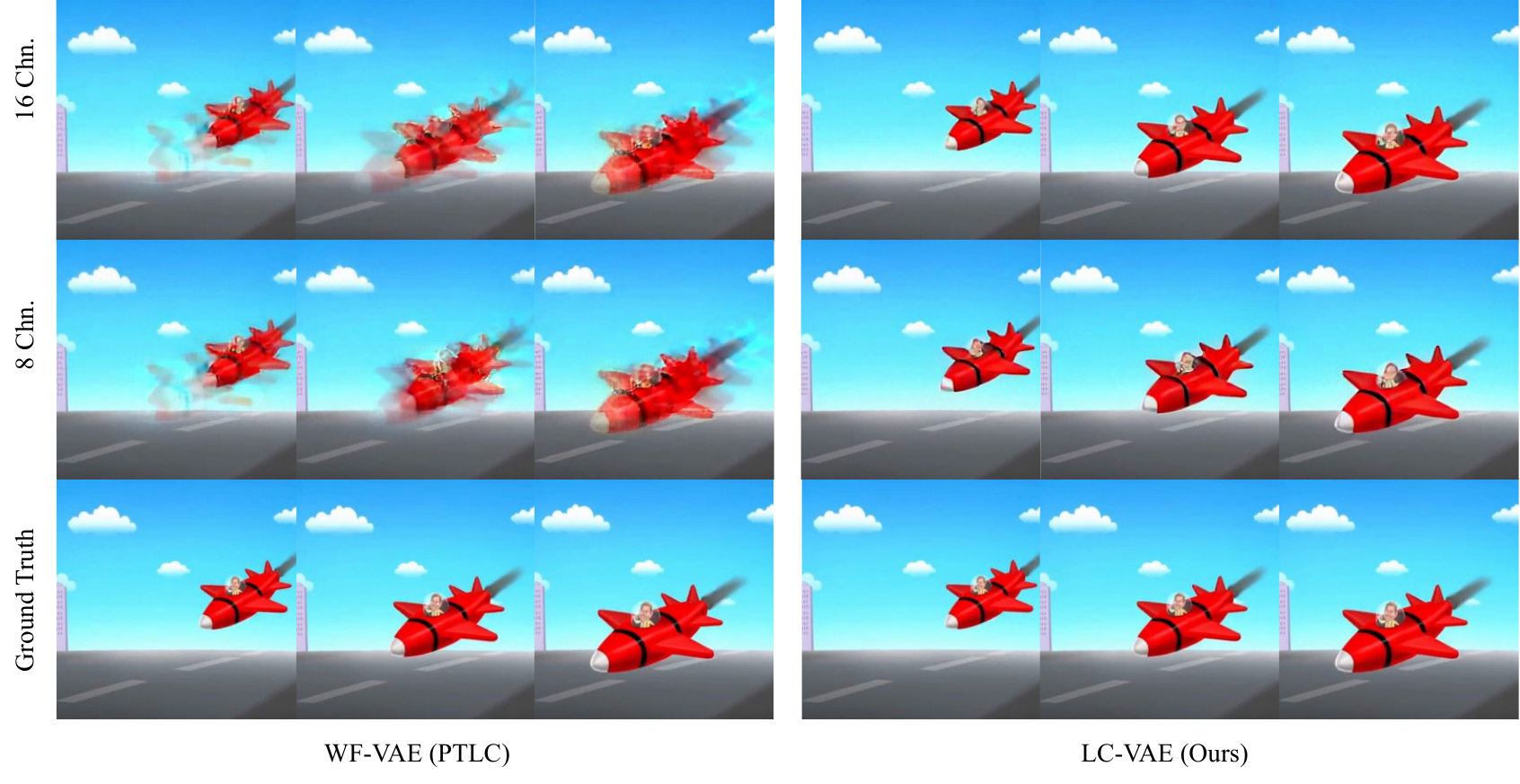} 
\caption{\textbf{Qualitative comparison between LC-VAE and WF-VAE (PTLC) at the same compression ratios (equivalent channels).} WF-VAE (PTLC) exhibits noticeable artifacts, whereas LC-VAE trained with latent compression reconstructs videos accurately, highlighting the importance of integrating latent compression during autoencoder training.}
\vspace{-2mm}
\label{fig:frequency_removal} 
\end{figure*}

\paragraph{Removing High-Frequency Components.} Building on the above observations, we examine whether the baseline model effectively offloads the reconstruction of fine details to the decoder. To this end, we conduct a high-frequency removal experiment on both raw videos and their corresponding latent representations. As shown in~\cref{fig:frequency_removal} and detailed in~\cref{sec:ablation}, removing high-frequency components from the WF-VAE latent space results in a substantial deterioration in reconstruction quality. This behavior contrasts with the effect of removing high-frequency components directly from the videos, where perceptual quality remains largely preserved~\cite{skorokhodov2025improving}. These findings suggest that the learned latents are overly sensitive to high-frequency information, indicating insufficient robustness in the baseline representation.

Taken together, these results indicate that existing video autoencoders without appropriate latent regularization may fail to prioritize the essential low-frequency structures of video content. A similar observation is reported by~\cite{skorokhodov2025improving}, where the authors demonstrate that increasing the number of channels in the autoencoder bottleneck amplifies high-frequency components, which become unevenly distributed and lack structural coherence across channels. Such unstructured latent representations not only reduce compression efficiency, as illustrated in~\cref{fig:multi_channel_comparison} where the reconstruction performance gain diminishes with increasing channel count, but also hinder diffusion model training, which inherently favors low-frequency data~\cite{chen2025dc}.

\begin{figure*}[t!]
    \centering\footnotesize
    %
    \includegraphics[width=0.85\textwidth, trim={0 0cm 0 0.0cm}, clip]{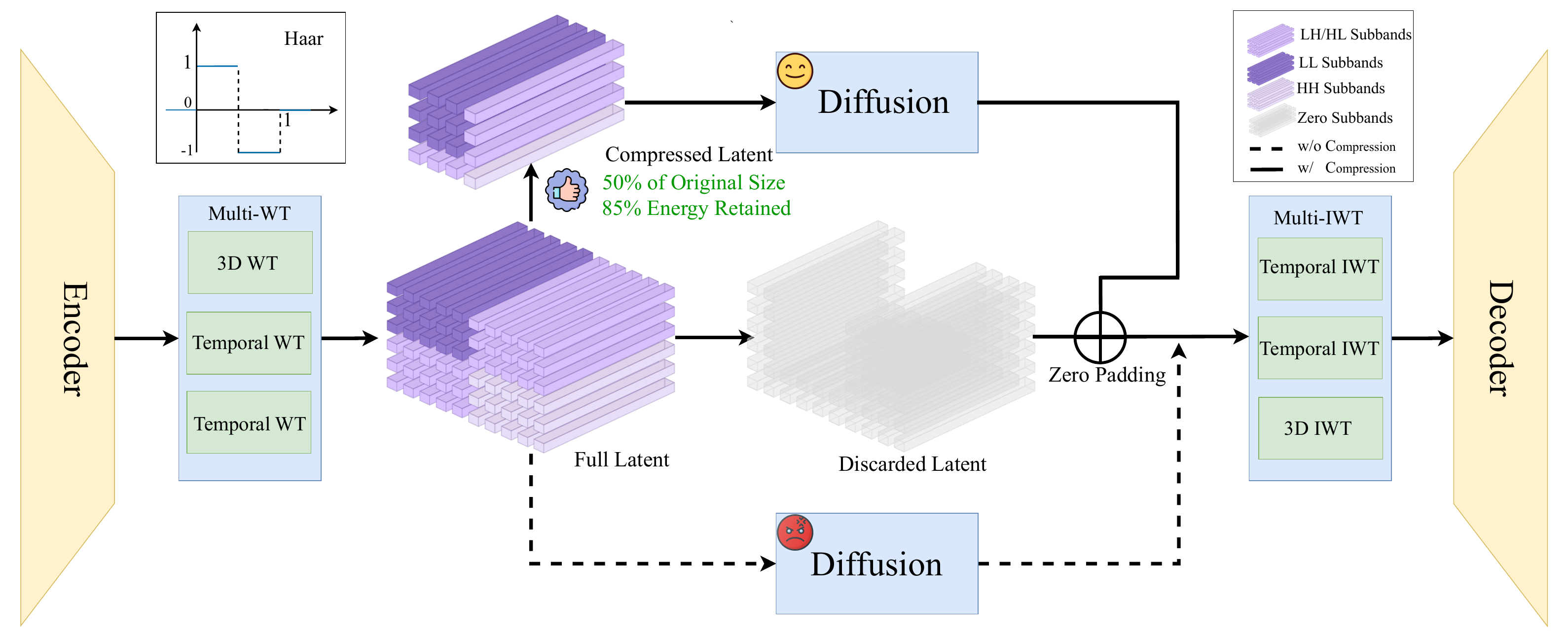}
    \caption{
    \textbf{Overview of our framework.} 
    The model first applies a multi-level wavelet transform (Multi-WT) to the latent features produced by the encoder. Low-frequency channels are then selected to retain compact yet informative representations in the wavelet domain, while the high-frequency subbands are zeroed out. During generation, diffusion operates within this favorable and compressed subspace. The sampled representation is subsequently zero-padded, passed through multi-scale inverse wavelet transforms (Multi-IWT), and finally decoded to reconstruct the video.
    }
    \label{fig:overview}
\end{figure*}

\subsection{Model Architecture}
\label{sec:architecture}


To address the limitations of existing video autoencoders, we propose compressing the video latent representation further, rather than directly reducing the number of latent channels, which often significantly degrades reconstruction fidelity. For a fair comparison, our autoencoder retains the same architecture as the main baseline model WF-VAE~\cite{li2025wf}, with the exception that we incorporate the newly proposed latent compression and reconstruction procedures described below.

Let the encoder map a video $\mathbf{v} \in \mathbb{R}^{C \times T \times H \times W}$ to a latent representation:
\begin{equation}
\mathbf{z} = \mathcal{E}(\mathbf{v}) \in \mathbb{R}^{C_z \times T_z \times H_z \times W_z}.
\end{equation}
Instead of reducing latent channels, we apply a multi-level 3D Haar wavelet transform to the latent:
\begin{equation}
\mathbf{B}^{(\ell)}_{abc} = \mathbf{B}^{(\ell-1)} \circledast (\xi_a \otimes \xi_b \otimes \xi_c), 
\quad \mathbf{B}^{(0)} = \mathbf{z},
\end{equation}
where $\xi_a, \xi_b, \xi_c \in \{\phi, \psi\}$ denote the low-pass and high-pass filters along temporal, height, and width dimensions, respectively, following the notation in \cref{equ:wavelet}. By decomposing the latent into subbands with distinct frequency characteristics, we selectively zero out the high-frequency subbands while retaining only the low-frequency subband for downstream learning:
\begin{equation}
\tilde{\mathbf{B}}^{(\ell)}_{abc} =
\begin{cases}
\mathbf{B}^{(\ell)}_{abc}, & \text{if } abc \in \{\textrm{LLL, LLH, LHL, HLL}\}\\
0, & \text{otherwise (high-frequency)}.
\end{cases}
\label{equ:freq_zero_out}
\end{equation}
Our fixed zero-out design is motivated by classical compression principles; we validate that it closely approximates a data-driven adaptive selection strategy in \cref{sec:adaptive_ablation}.
Before decoding, the filtered latent is padded with zero-valued high-frequency subbands and transformed back via the inverse wavelet transform:
\begin{equation}
\tilde{\mathbf{z}} = \mathcal{W}^{-1}\big(\{\tilde{\mathbf{B}}^{(\ell)}_{abc}\}\big),
\end{equation}
from which the decoder reconstructs the video:
\begin{equation}
\tilde{\mathbf{v}} = \mathcal{D}(\tilde{\mathbf{z}}).
\end{equation}
This design forces the encoder to focus exclusively on diffusion-favorable and compression-efficient low-frequency information, while delegating the reconstruction of high-frequency details to the decoder.

By performing multi-level wavelet filtering in the latent space, our LC-VAE preserves essential information for downstream diffusion models while reducing the dimensionality and redundancy of the latent representation.

\subsection{Training}
\label{sec:training}
Our loss function follows the training strategies of existing works~\cite{rombach2022high, esser2021taming} and is set the same as WF-VAE~\cite{li2025wf}. Specifically, the final loss function is formulated as follows,
\begin{equation}
    \mathcal{L}=\mathcal{L}_{\textrm{recon}}+\lambda_{\textrm{adv}}\mathcal{L}_{\textrm{adv}}+\lambda_{\textrm{KL}}\mathcal{L}_{\textrm{KL}},
    \label{equ:loss}
\end{equation}
where $\mathcal{L}_{\textrm{recon}}, \mathcal{L}_{\textrm{adv}},$ and $\mathcal{L}_{\textrm{KL}}$ represent the $\textrm{L1}$ reconstruction loss, adversarial loss~\cite{goodfellow2014generative}, $\textrm{KL}$ regularization loss. $\lambda$ represents the corresponding weight for each loss item.



\section{Experiments}
We design a set of experiments to (1) extensively examine the video compression and reconstruction performance of the proposed LC-VAE, (2) examine whether the low-frequency-focused latents in LC-VAE can exhibit improved generalization ability compared to the baseline method, (3) assess the necessity of jointly training the autoencoder with latent compression, and (4) demonstrate the video generation results.


\subsection{Experimental Setup}
\paragraph{Baseline Models.} 
We conduct a comprehensive evaluation of LC-VAE by comparing it against several state-of-the-art video VAE models. Our baseline suite includes: (1) WF-VAE~\cite{li2025wf}, a 3D causal convolutional architecture employed in Open-Sora Plan 1.5~\cite{lin2024open}; (2) Open-Sora VAE~\cite{zheng2024open}; (3) CV-VAE~\cite{zhao2024cv}; (4) OD-VAE~\cite{chen2025od} used in Open-Sora Plan 1.2~\cite{lin2024open}; (5) SVD-VAE~\cite{blattmann2023stable}, which operates without temporal compression; and (6) SD-VAE~\cite{rombach2022high}, a widely-adopted image VAE baseline. 

\paragraph{Datasets and Evaluation Metrics.} 
We utilize the Kinetics-400 dataset~\cite{kay2017kinetics} for model training and validation. For evaluation, we perform zero-shot testing on the Panda-70M~\cite{chen2024panda} and WebVid-10M~\cite{bain2021frozen} datasets. 

To comprehensively assess reconstruction quality, we adopt multiple complementary metrics: Peak Signal-to-Noise Ratio (PSNR)~\cite{hore2010image} for pixel-level fidelity, Learned Perceptual Image Patch Similarity (LPIPS)~\cite{zhang2018unreasonable} for perceptual quality, and Structural Similarity Index Measure (SSIM)~\cite{wang2004image} for structural preservation. Additionally, we employ reconstruction Fréchet Video Distance (rFVD)~\cite{unterthiner2019fvd} to evaluate visual quality and temporal coherence.

\begin{table*}[ht]
 \centering 
 \caption{\textbf{Quantitative metrics of reconstruction performance.} Results demonstrate that LC-VAE achieves state-of-the-art on reconstruction performance comparing with other VAEs on WebVid-10M~\cite{bain2021frozen} and Panda-70M~\cite{chen2024panda} datasets. TCPR represents the token compression rate, and Chn. indicates the number of latent channels. The best result is highlighted in \textbf{bold}.}
 \label{tab:table_main_result}
\resizebox{\textwidth}{!}{%
\begin{tabular}{c| cc cccc cccc}
 \toprule[1.2pt]
   \multirow{2}{*}{\textbf{Method}}  & \multirow{2}{*}{\textbf{TCPR}} & \multirow{2}{*}{\textbf{Chn.}} & \multicolumn{4}{c}{\textbf{WebVid-10M}} & \multicolumn{4}{c}{\textbf{Panda-70M}}\\
   \cmidrule(lr){4-7} \cmidrule(lr){8-11}
   &&&  \textbf{PSNR} $(\uparrow)$ & \textbf{SSIM}$(\uparrow)$ & \textbf{LPIPS} $(\downarrow)$ & \textbf{rFVD} $(\downarrow)$ & \textbf{PSNR} $(\uparrow)$ & \textbf{SSIM} $(\uparrow)$ & \textbf{LPIPS}$(\downarrow)$ & \textbf{rFVD} $(\downarrow)$\\
\midrule
SD-VAE~\cite{rombach2022high}&$64(1\times8\times8)$&4&30.19&0.8377&0.0568&284.90 &30.46&0.8896&0.0395&182.99\\
SVD-VAE~\cite{blattmann2023stable}&$64(1\times8\times8)$&4&31.18&0.8689&0.0546&188.74&31.04&0.9059&0.0379&137.67\\
\midrule
CV-VAE~\cite{zhao2024cv}&$256(4\times8\times8)$&4&30.76&0.8566&0.0803&369.23 &30.18&0.8796&0.0672&296.28\\
OD-VAE~\cite{chen2025od}&$256(4\times8\times8)$&4&30.69&0.8635&0.0553&255.92&30.31&0.8935&0.0439&\textbf{191.23}\\
Open-Sora VAE ~\cite{zheng2024open}&$256(4\times8\times8)$&4&31.14&0.8572&0.1001&475.23&31.37&\textbf{0.8973}&0\textbf{.}0662&298.47\\
WF-VAE~\cite{li2025wf} &$256(4\times8\times8)$&4&30.68&0.9071&0.0344&179.13&31.22&0.8713&0.0474&261.89\\
LC-VAE (Ours)&$256(4\times8\times8)$&4&\textbf{31.49}&\textbf{0.9207}&\textbf{0.0249}&\textbf{165.88}&\textbf{31.89}&0.8909&\textbf{0.0303}&225.98\\
\midrule
WF-VAE~\cite{li2025wf} &$256(4\times8\times8)$&8&31.96&\textbf{0.9281}&0.0242&\textbf{101.06}&32.41&0.8982&0.0348&156.95\\
LC-VAE (Ours)&$256(4\times8\times8)$&8&\textbf{33.78}&0.9208&\textbf{0.0211}&135.99&\textbf{33.64}&\textbf{0.9447}&\textbf{0.0165}&\textbf{93.89}\\
\midrule
WF-VAE~\cite{li2025wf} &$256(4\times8\times8)$&16&34.62&0.9301&0.0193&\textbf{68.72}&34.68&0.9542&0.0133&\textbf{42.94} \\
LC-VAE (Ours)&$256(4\times8\times8)$&16&\textbf{35.59}&\textbf{0.9439}&\textbf{0.0152}&73.66&\textbf{35.63}&\textbf{0.9616}&\textbf{0.0032}&50.40\\
\bottomrule[1.2pt]
\end{tabular}}

\end {table*}

\begin{table*}[h!]
\centering
\caption{
\textbf{Zero-shot reconstruction comparison across three datasets.}
Our LC-VAE maintains consistently high performance across all datasets with smaller degradations, while the baseline WF-VAE suffers a noticeable performance drop of $0.5$--$1.5$ dB in PSNR when tested on unseen datasets.}
\label{tab:zero_shot}
\resizebox{\textwidth}{!}{
\begin{tabular}{c|cc|cccc|cccc|cccc}
\toprule[1.2pt]
\multirow{2}{*}{\textbf{Method}} & \multirow{2}{*}{\textbf{TCPR}} & \multirow{2}{*}{\textbf{Chn.}} &
\multicolumn{4}{c|}{\textbf{UCF-101}} & 
\multicolumn{4}{c|}{\textbf{SkyTimelapse}} &
\multicolumn{4}{c}{\textbf{OpenVid-1M}} \\ 
\cmidrule(lr){4-7}\cmidrule(lr){8-11}\cmidrule(lr){12-15}
&&& \textbf{PSNR}$\uparrow$ & \textbf{SSIM}$\uparrow$ & \textbf{LPIPS}$\downarrow$ & \textbf{rFVD}$\downarrow$ &
\textbf{PSNR}$\uparrow$ & \textbf{SSIM}$\uparrow$ & \textbf{LPIPS}$\downarrow$ & \textbf{rFVD}$\downarrow$ &
\textbf{PSNR}$\uparrow$ & \textbf{SSIM}$\uparrow$ & \textbf{LPIPS}$\downarrow$ & \textbf{rFVD}$\downarrow$ \\
\midrule
WF-VAE & $256(4\times8\times8)$ & 4 & 30.32 &0.8955 &0.0431 & \textbf{337.69} & 36.59 & 0.9444 & 0.0199 & 100.71& 32.98 &0.8831 & 0.0418 & 193.08 \\
LC-VAE (Ours) & $256(4\times8\times8)$ & 4 & \textbf{30.57} & \textbf{0.9005} & \textbf{0.0333} & 340.17 & \textbf{36.87} & \textbf{0.9489} & \textbf{0.0146} & \textbf{87.06} & \textbf{33.48} & \textbf{0.8993} & \textbf{0.0242} & \textbf{145.04} \\
\midrule
WF-VAE & $256(4\times8\times8)$ & 8 & 31.86& 0.9204 &0.0294&\textbf{189.54} & 37.71 & 0.9541 & 0.0139&61.71 & 34.20 & 0.9066 & 0.0311 & 109.89 \\
LC-VAE (Ours) & $256(4\times8\times8)$ & 8 & \textbf{32.69} &\textbf{0.9292} & \textbf{0.0231} & 198.94 & \textbf{38.65} &\textbf{0.9620}  & \textbf{0.0103}& \textbf{50.43} & \textbf{35.27} & \textbf{0.9253} & \textbf{0.0172} & \textbf{89.38} \\
\midrule
WF-VAE & $256(4\times8\times8)$ & 16 & 34.40 & 0.9488 &  0.0171 & \textbf{81.29} & 39.85 & 0.9681 &0.0084 &28.06 & 36.28& 0.9332 &  0.0167 & 50.96 \\
LC-VAE (Ours) & $256(4\times8\times8)$ & 16 & \textbf{34.83} & \textbf{0.9514} & \textbf{0.0159} &106.09 &\textbf{40.30}  & \textbf{0.9718} &\textbf{0.0076} & \textbf{27.31}& \textbf{37.06} & \textbf{0.9463} & \textbf{0.0123} & \textbf{50.41} \\
\bottomrule[1.2pt]
\end{tabular}}
\end{table*}

To assess generative capabilities when integrated with diffusion models, we conduct experiments on the UCF-101~\cite{soomro2012ucf101} and SkyTimelapse~\cite{xiong2018learning} datasets for conditional and unconditional generation tasks, respectively. Following established protocols~\cite{skorokhodov2022stylegan, ma2025latte, li2025wf}, we extract 16-frame clips from 2,048 videos to compute FVD$_{16}$ scores. We also evaluate Inception Score (IS)~\cite{saito2017temporal} exclusively on UCF-101, as in ~\cite{ma2025latte}. For the denoising model, we select Latte-L~\cite{ma2025latte}, training for 100,000 steps. Note that our focus is on evaluating whether the latent spaces of different video VAEs facilitate effective diffusion model training rather than maximizing generative performance.

\paragraph{Training Configuration.} 
We employ the AdamW optimizer~\cite{adam2014method, loshchilov2017decoupled} with $\beta_1 = 0.9$, $\beta_2 = 0.999$, and a fixed learning rate of $1 \times 10^{-5}$. Different from WF-VAE~\cite{li2025wf} that has a sophisticated three-stage training, we simply apply one stage training with fixed loss function and hyper-parameters. We train our model and baseline WF-VAE model with the same setting for 200,000 steps. Our training is conducted on 8 NVIDIA H200 GPUs for around 5 days. Complete hyperparameter specifications are provided in the appendix.

\subsection{Main Results}
We evaluate the performance of LC-VAE against baseline methods in terms of both video reconstruction quality and diffusion-based generation capability. We further compare three variants of our model with different numbers of latent channels to the original WF-VAE. It is worth noting that our latent compression approach is model-agnostic and can be readily applied to other VAE backbones. We verify this by applying LC-VAE to WanVAE-2.1 and scaling the diffusion
model to 2.1B parameters in \cref{sec:scalability}.

\begin{figure}[ht]
  \centering
  \includegraphics[width=0.96\linewidth, trim={0 0cm 0 0.0cm}, clip]{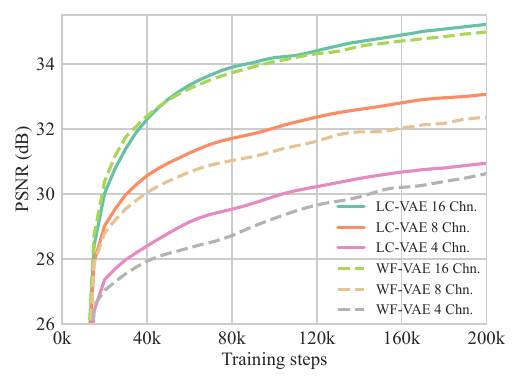}
\caption{\textbf{Validation performance during training.} Across different compression ratios (Chn. = 4, 8, 16), our method consistently achieves higher PSNR than the baseline.}
\vspace{-2mm}
\label{fig:training_plot}
\end{figure}

\begin{figure*}[t!]
    \centering\footnotesize
    \vspace*{-0.8em}
    \includegraphics[width=0.98\textwidth, trim={0 0cm 0 0.2cm}, clip]{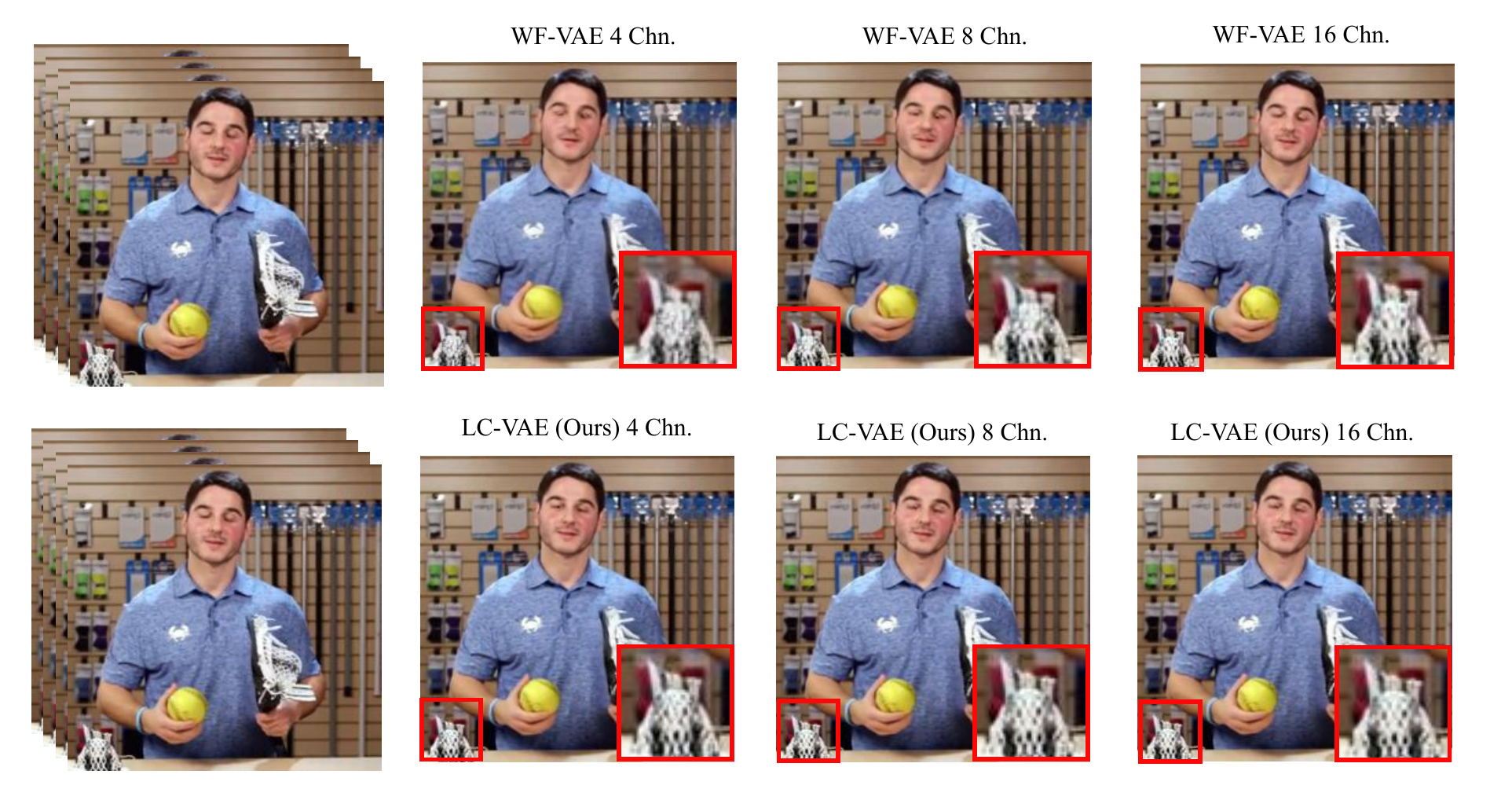}

    \caption{\textbf{Qualitative comparison of reconstruction performance between LC-VAE and WF-VAE under the same compression ratios (equivalent channels).}}
    \label{fig:qualitative_comparison}
\end{figure*}

\paragraph{Video Reconstruction Evaluation.}
We report quantitative results in~\cref{tab:table_main_result} and qualitative comparisons in~\cref{fig:qualitative_comparison}. The validation performance throughout training is shown in~\cref{fig:training_plot}. As observed in~\cref{tab:table_main_result}, LC-VAE consistently achieves the highest PSNR with clear margins and generally outperforms other methods across other evaluation metrics. For instance, on the WebVid-10M dataset, LC-VAE surpasses the WF-VAE counterparts by 0.81 dB, 1.82 dB, and 0.97 dB of PSNR for 4, 8, and 16 latent channels, respectively. The qualitative results in~\cref{fig:qualitative_comparison} further confirm the enhanced perceptual quality achieved by LC-VAE.

\begin{table}[h!]
\centering
\caption{\textbf{Quantitative evaluation for video generation.} We compare WF-VAE and the proposed LC-VAE across different channel sizes on SkyTimelapse and UCF-101 datasets. 
}
\label{tab:FVDResults}
\setlength{\tabcolsep}{12pt}
\resizebox{\columnwidth}{!}
{
\begin{tabular}{l c c cc}
\toprule
\multirow{2}{*}{\textbf{Method}} & \multirow{2}{*}{\textbf{Chn.}} 
& \multicolumn{1}{c}{\textbf{SkyTimelapse}} 
& \multicolumn{2}{c}{\textbf{UCF-101}} \\
\cmidrule(lr){3-3}\cmidrule(lr){4-5}
& & \textbf{FVD$_{16}$} $\downarrow$ & \textbf{FVD$_{16}$} $\downarrow$ & \textbf{IS} $\uparrow$ \\
\midrule
WF-VAE                                   & 4 & \textbf{198.87} & 565.80 & 61.19 \\
LC-VAE (Ours)                            & 4 & 240.56 & \textbf{ 509.76
} & \textbf{70.71 } \\
\midrule
WF-VAE                                   & 8 & 213.23& 687.60 &60.57  \\
LC-VAE (Ours)                            & 8 &  \textbf{201.24}&  \textbf{654.96} & \textbf{66.72}\\
\midrule
WF-VAE                                   & 16 & 195.94 & \textbf{721.43} & 52.66  \\
LC-VAE (Ours)                            & 16 & \textbf{187.68} & 735.04 & \textbf{54.89} \\

\bottomrule
\end{tabular}}
\end{table}

\begin{figure}[ht]
  \centering
  \includegraphics[width=0.96\linewidth, trim={0 0 0 0}, clip]{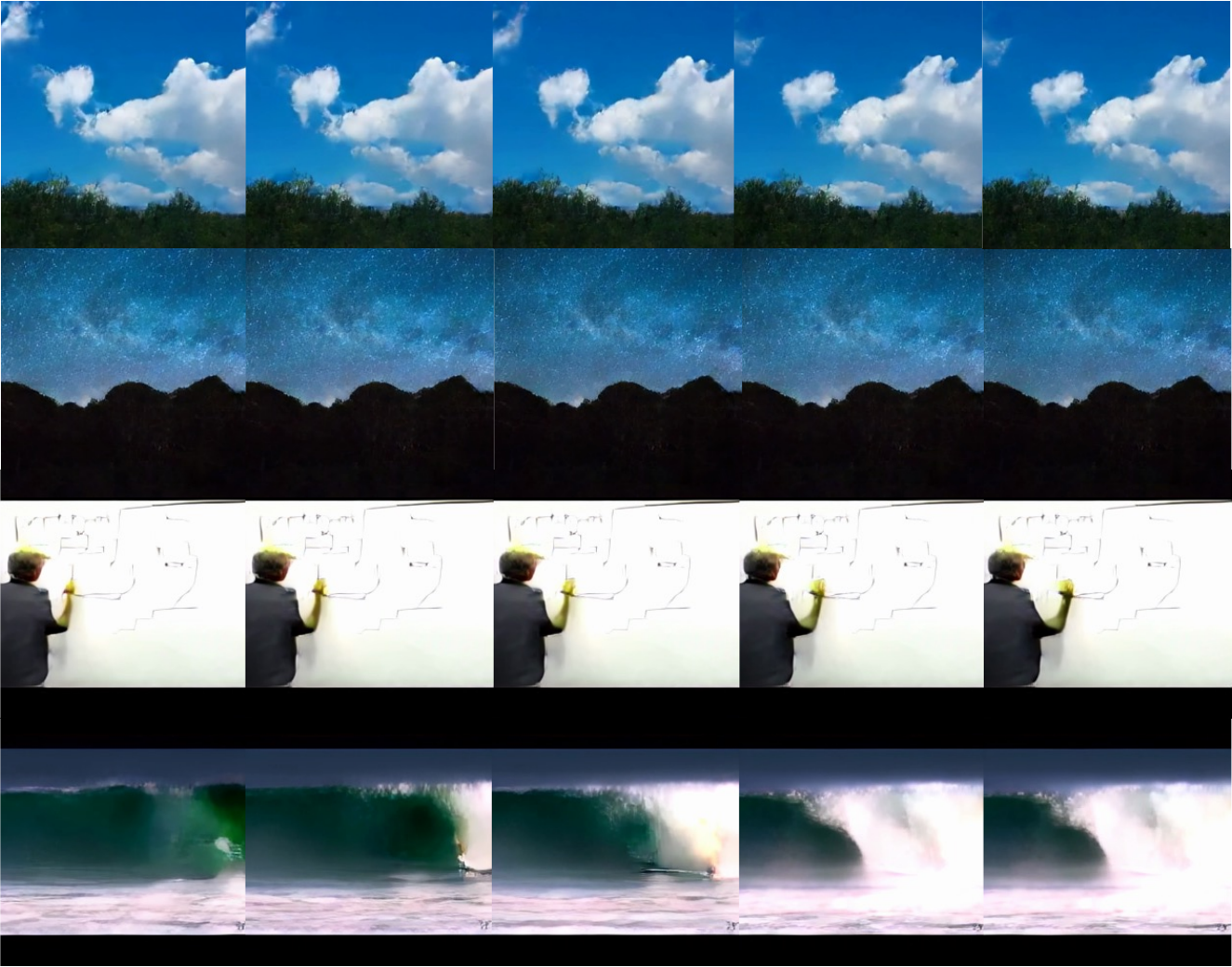}\\
  \caption{\textbf{Generated videos using LC-VAE with Latte~\cite{ma2025latte} on SkyTimelapse (top) and UCF-101 (bottom) datasets}.}
  \label{fig:gen_videos}
  \vspace{-6mm}
\end{figure}

\paragraph{Additional Zero-shot Reconstruction Comparison.} As shown in~\cref{fig:training_plot}, the baseline WF-VAE achieves comparable validation performance on the training Kinetics-400 dataset. However, on the zero-shot test datasets reported in~\cref{tab:table_main_result}, our method clearly outperforms the baseline. We hypothesize that enforcing the encoder to focus on learning robust low-frequency representations improves its generalization ability and thus benefits zero-shot transfer. To further evaluate this property, we assess the zero-shot reconstruction performance on three additional datasets: two used for video generation training, UCF-101~\cite{soomro2012ucf101} and SkyTimelapse~\cite{xiong2018learning}, as well as the newly introduced OpenVid~\cite{nan2024openvid}. The comparison results in~\cref{tab:zero_shot} show that LC-VAE consistently achieves higher PSNR scores than WF-VAE with clear margins across all datasets. 

\paragraph{Video Generation Evaluation.} We present both quantitative and qualitative results for video generation via diffusion using LC-VAE in~\cref{tab:FVDResults} and~\cref{fig:gen_videos}. For all experiments, we adopt the same training configuration as the baseline WF-VAE. The results indicate that LC-VAE generally achieves lower FVD scores across models with varying channel sizes and datasets. Notably, diffusion with WF-VAE (4 channels) attains better performance than LC-VAE(4 channels) on the Sky Timelapse dataset. We hypothesize that this dataset predominantly consists of static videos, containing limited high-frequency components. Consequently, the baseline model performs well with compact latent representations.

\begin{table}[ht]
\centering
\caption{\textbf{Ablation on the necessity of latent compression during training.}
We perform post-training latent compression (PTLC) on WF-VAE and compare its performance with LC-VAE across same channel settings, Chn.$\in{8,16}$, on the WebVid-10M and Panda-70M datasets. LC-VAE consistently achieves superior reconstruction quality on both datasets.}
\footnotesize
\setlength{\tabcolsep}{3.5pt}
\renewcommand{\arraystretch}{1.12}
\resizebox{\columnwidth}{!}{
\begin{tabular}{l c ccc ccc}
\toprule
\multirow{2}{*}{\textbf{Method}} & \multirow{2}{*}{\textbf{Chn.}} 
& \multicolumn{3}{c}{\textbf{WebVid-10M}} 
& \multicolumn{3}{c}{\textbf{Panda-70M}} \\
\cmidrule(lr){3-5} \cmidrule(lr){6-8}
& & \textbf{PSNR}$\uparrow$ & \textbf{SSIM}$\uparrow$ & \textbf{LPIPS}$\downarrow$
  & \textbf{PSNR}$\uparrow$ & \textbf{SSIM}$\uparrow$ & \textbf{LPIPS}$\downarrow$ \\
\midrule
WF-VAE (PTLC) & 8  & 29.24 & 0.8393 & 0.0675  &  27.51  & 0.8526 &0.0781  \\
LC-VAE        & 8  & \textbf{31.49} & \textbf{0.9207} & \textbf{0.0249}  
               & \textbf{31.89} & \textbf{0.8909} & \textbf{0.0303} \\
\midrule
WF-VAE (PTLC) & 16 & 30.49 & 0.8725 & 0.0545  &  28.67& 0.8744 &0.0684  \\
LC-VAE        & 16 & \textbf{33.78} & \textbf{0.9208} & \textbf{0.0211}  
               & \textbf{33.64} & \textbf{0.9447} & \textbf{0.0165} \\
\bottomrule
\end{tabular}}
\vspace{-3mm}
\label{tab:ablate_keep_ratio_dual}
\end{table}



\subsection{Ablation Study}
\label{sec:ablation}
Our method is straightforward to implement and does not introduce additional hyperparameters. 
LC-VAE is characterized by performing latent compression and reconstruction during training, thereby delegating the recovery of high-frequency details to the decoder.
We hypothesize that directly applying latent compression to a pre-trained video VAE would result in a noticeable performance degradation, as the latent representation contains unstructured high-frequency information and the decoder has not been explicitly trained to reconstruct such details.

\paragraph{Post-training Latent Compression.} We compare our method with a new baseline that applies \textit{post-training latent compression} (PTLC) to pre-trained WF-VAE models. Specifically, we perform the same three-level wavelet transform on the WF-VAE latent representations and zero out the high-frequency subbands, following the procedure used in our LC-VAE. PTLC is applied to WF-VAE models with 16 and 8 latent channels, resulting in compressed representations equivalent to 8 and 4 channels, respectively. We then compare the reconstruction performance with LC-VAE under identical compression ratios. As shown in~\cref{tab:ablate_keep_ratio_dual}, our method consistently outperforms the PTLC baseline, demonstrating that integrating latent compression into the autoencoder training process is crucial for achieving high reconstruction quality.

\section{Conclusion}

We have shown that existing video VAEs typically learn latent representations with unstructured frequency distributions. As reported in prior work, these latents often contain uninformative~\cite{chen2025dc} or even detrimental~\cite{skorokhodov2025improving} components due to the presence of excessive high-frequency information. This problem becomes more pronounced as the number of latent channels increases. To address this issue, we propose a novel latent compression approach that removes high-frequency components from the video latent space. Extensive experiments across multiple datasets demonstrate that the proposed LC-VAE outperforms strong baselines under equivalent compression ratios.

\paragraph{Limitations.} As the first attempt to compress the latent space of video VAEs, our method simply zeros out high-frequency subbands without tuning the compression ratios. In addition, we do not aim to maximize video generation performance and therefore do not modify or optimize the downstream diffusion model architecture. Nevertheless, we hypothesize that LC-VAE learns video latents with reduced high-frequency noise, which could provide broader benefits to various video diffusion models.
\section*{Acknowledgments}

AS and JK acknowledge funding from the Research Council of Finland (352788, 362407, 373999, 373780, 362408, and 339730). This work was supported by the Research Council of Finland Flagship programme: Finnish Center for Artificial Intelligence FCAI. We acknowledge the computational resources provided by the Aalto Science-IT project.


{
    \small
    \bibliographystyle{ieeenat_fullname}
    \bibliography{main}
}

\clearpage
\setcounter{page}{1}
\maketitlesupplementary

\setcounter{section}{0}
\renewcommand{\thesection}{\Alph{section}}

\startcontents[supplement]
\printcontents[supplement]{}{1}{\subsection*{Contents}}

\section{Training Details}
\label{sec:training_details}

The training hyperparameters are listed in \cref{tb:train_hyper}.
A reference implementation and pretrained checkpoints can be found at: \url{https://github.com/1mather/LC-VAE-code}.

\begin{table}[h]
\centering
\caption{Training hyperparameters used in all experiments.}
\label{tb:train_hyper}
\renewcommand{\arraystretch}{1.15}
\setlength{\tabcolsep}{6pt}
\begin{tabular}{l c}
\toprule[1.2pt]
\textbf{Hyperparameter} & \textbf{Value} \\
\midrule
Training steps                           & 200k \\
Learning rate                            & $1\times10^{-5}$ \\
Total batch size                         & 32 \\
Perceptual (LPIPS) weight                & 1.0 \\
Adversarial loss weight ($\lambda_{\text{adv}}$) & dynamic \\
KL weight ($\lambda_{\text{KL}}$)        & $1\times10^{-6}$ \\
Resolution                               & $256\times256$ \\
Number of frames                         & 32 \\
EMA decay                                & 0.999 \\
\bottomrule[1.2pt]
\end{tabular}
\end{table}

Following~\cite{li2025wf}, we adopt a dynamic adversarial-loss weighting scheme
that balances the gradient magnitudes of the adversarial and reconstruction
objectives:
\begin{equation}
\lambda_{\text{adv}}
= \frac{1}{2}
\frac{\left\lVert \nabla_{G_L}\mathcal{L}_{\text{recon}} \right\rVert}
     {\left\lVert \nabla_{G_L}\mathcal{L}_{\text{adv}}   \right\rVert + \delta},
\end{equation}
where $\nabla_{G_L}(\cdot)$ denotes the gradient with respect to the
decoder's final layer and $\delta\!=\!10^{-6}$ ensures numerical stability.

\section{Multi-Level Wavelet Transform of Latent}
\label{sec:multi_wt}

\begin{figure}[t]
    \centering\footnotesize
    \includegraphics[width=\linewidth]{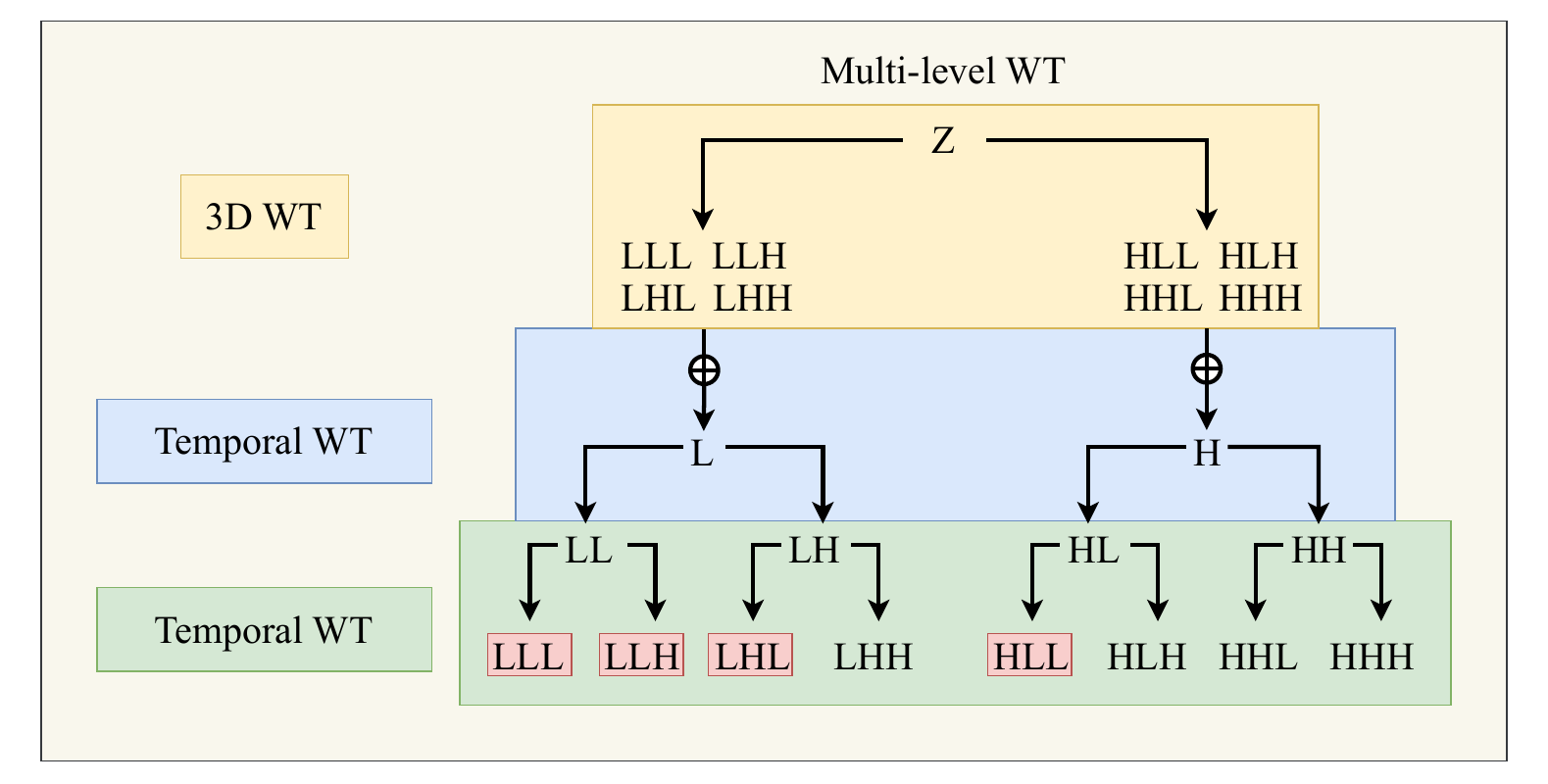}
    \caption{\textbf{Illustration of the proposed Multi-WT.}
    A 3D WT is first applied to the latent $\mathbf{z}$ to obtain eight
    subbands; two successive Temporal WT stages then further decompose them.
    In the Multi-WT representation the three letters (e.g., LHL) index
    temporal decomposition stages rather than spatial axes as
    in~\cref{equ:freq_zero_out}.
    We retain only the low-frequency–dominant subbands
    (\fcolorbox{red}{red!20}{LLL}, \fcolorbox{red}{red!20}{LLH},
     \fcolorbox{red}{red!20}{LHL}, \fcolorbox{red}{red!20}{HLL})
    and zero out the rest; $\oplus$ denotes channel-wise concatenation.}
    \label{fig:multi_level_wt}
    \vspace{-6mm}
\end{figure}

The Multi-WT proceeds in three stages.
\textbf{(1) 3D WT:} starting from $\mathbf{z}$, a 3D WT along the temporal,
height, and width dimensions yields eight subbands LLL, LLH, LHL, LHH, HLL,
HLH, HHL, HHH, where each character (L/H) denotes a low-/high-frequency
component along one axis.
\textbf{(2) Second-level Temporal WT:} the eight subbands are grouped by their
temporal component and a Temporal WT is applied to each group, producing four
subbands LL, LH, HL, HH.
\textbf{(3) Third-level Temporal WT:} a further Temporal WT on each of the four
subbands yields eight subbands again denoted LLL–HHH.

This Multi-WT notation differs from the basic 3D WT
in~\cref{equ:basic_wavelet_subbands}: there, the three letters index the
temporal, height, and width axes; here they index the first-, second-, and
third-stage decompositions.
We empirically select LLL, LLH, LHL, HLL as the low-frequency–dominant
subbands to retain, guided by the energy and autocorrelation analyses
in \cref{sec:freq_analysis}.

\section{Latent Frequency Analysis}
\label{sec:freq_analysis}

We provide a comprehensive characterisation of the latent frequency spectrum
through visualisation (\cref{sec:vis}), energy analysis (\cref{sec:energy}),
and temporal autocorrelation (\cref{sec:autocorr}).

\subsection{Frequency Subband Visualisation}
\label{sec:vis}

\begin{figure*}[t]
    \centering
    \begin{subfigure}[b]{0.49\textwidth}
        \centering
        \includegraphics[width=\linewidth]{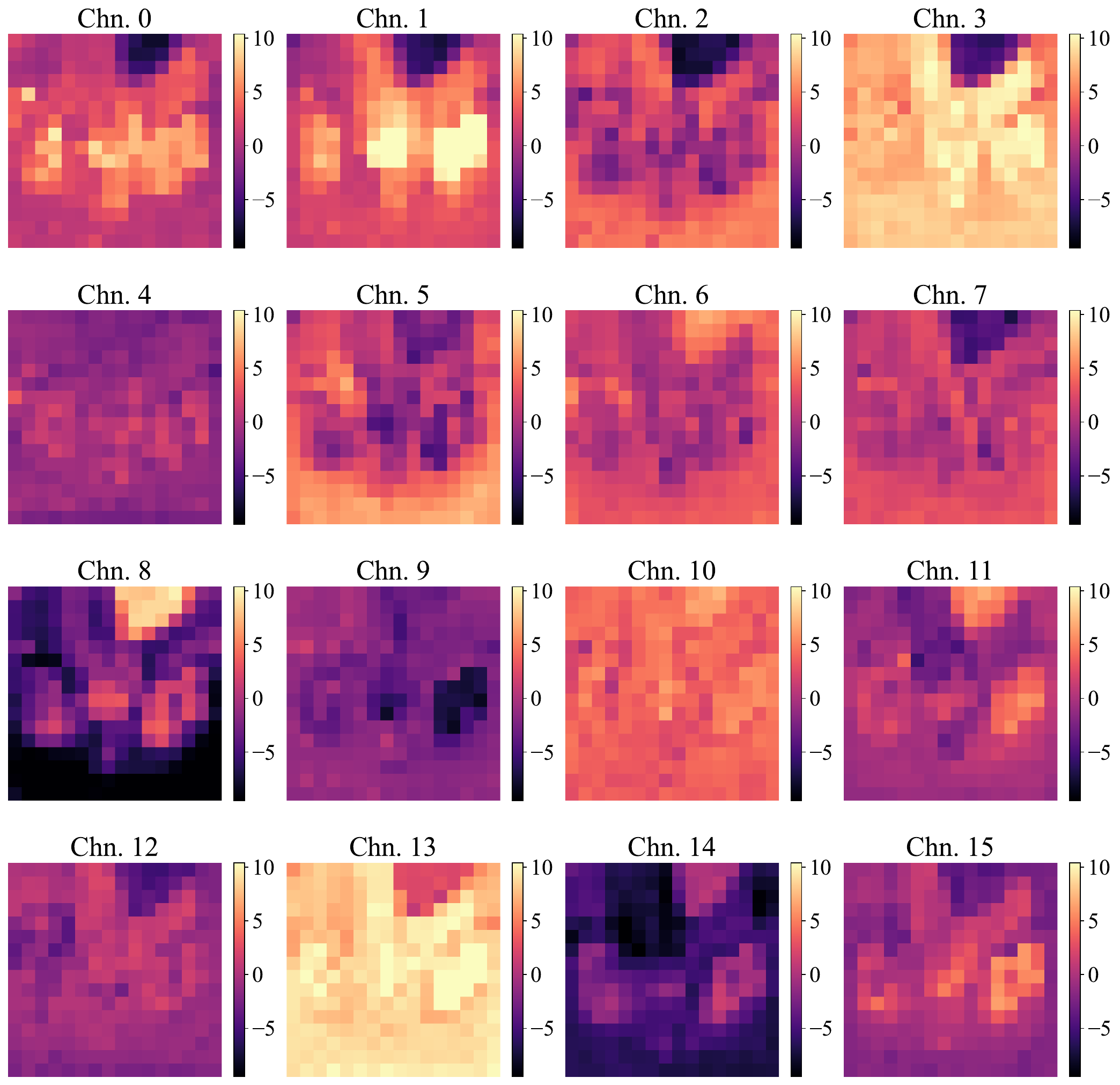}
        \caption{LLL}
    \end{subfigure}
    \hfill
    \begin{subfigure}[b]{0.49\textwidth}
        \centering
        \includegraphics[width=\linewidth]{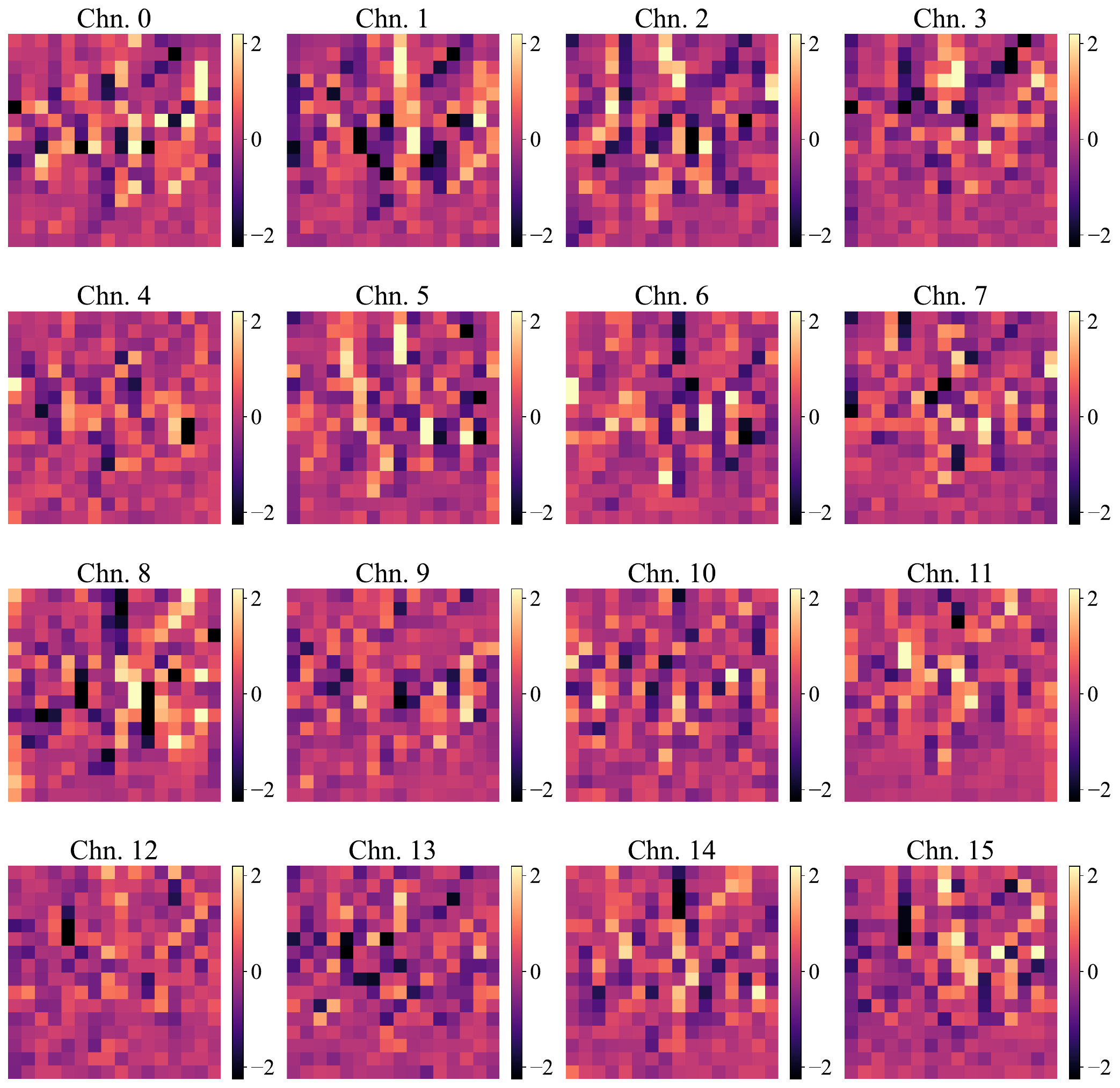}
        \caption{LLH}
    \end{subfigure}
    \vspace{0.6em}
    \begin{subfigure}[b]{0.49\textwidth}
        \centering
        \includegraphics[width=\linewidth]{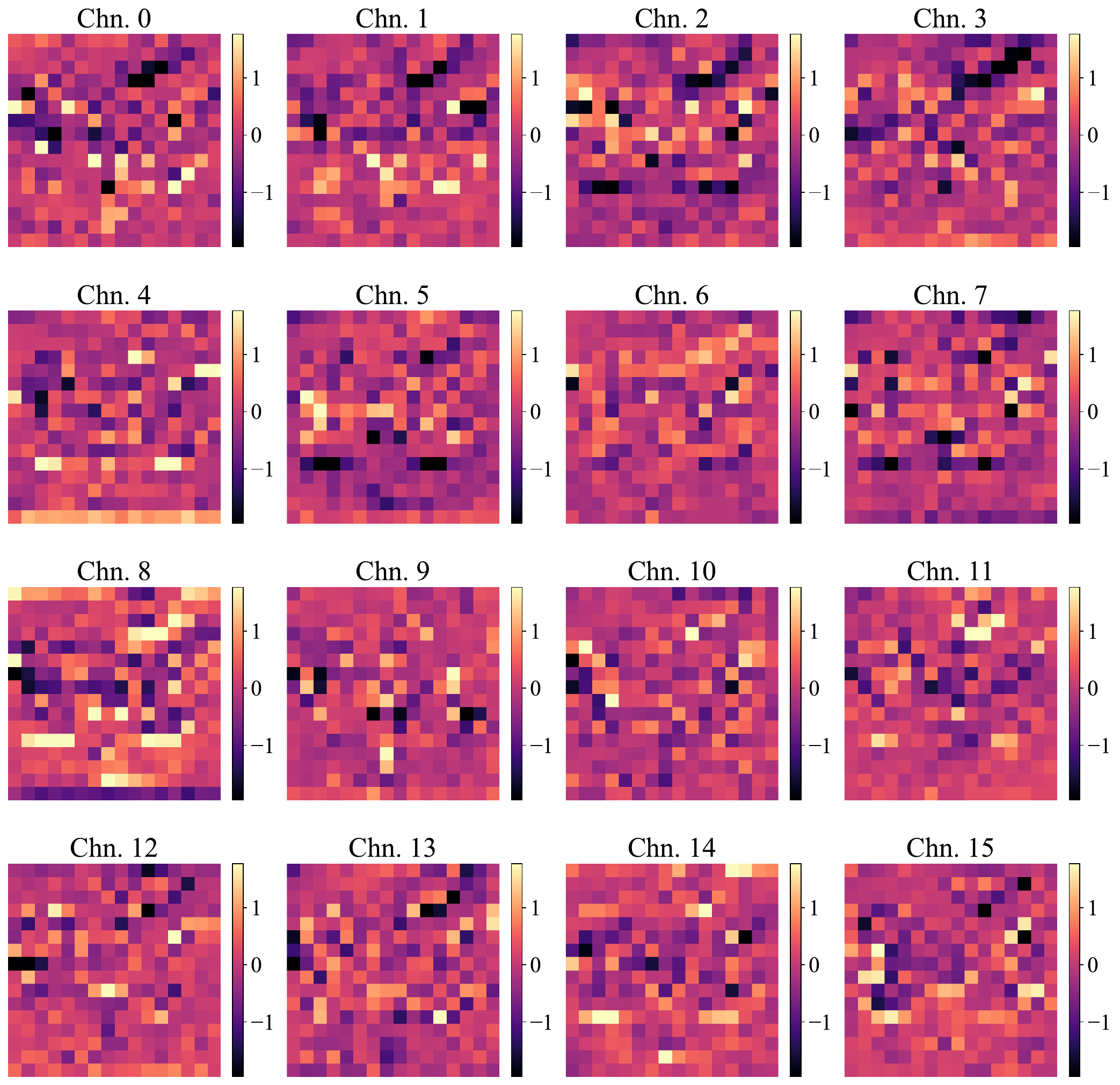}
        \caption{LHL}
    \end{subfigure}
    \hfill
    \begin{subfigure}[b]{0.49\textwidth}
        \centering
        \includegraphics[width=\linewidth]{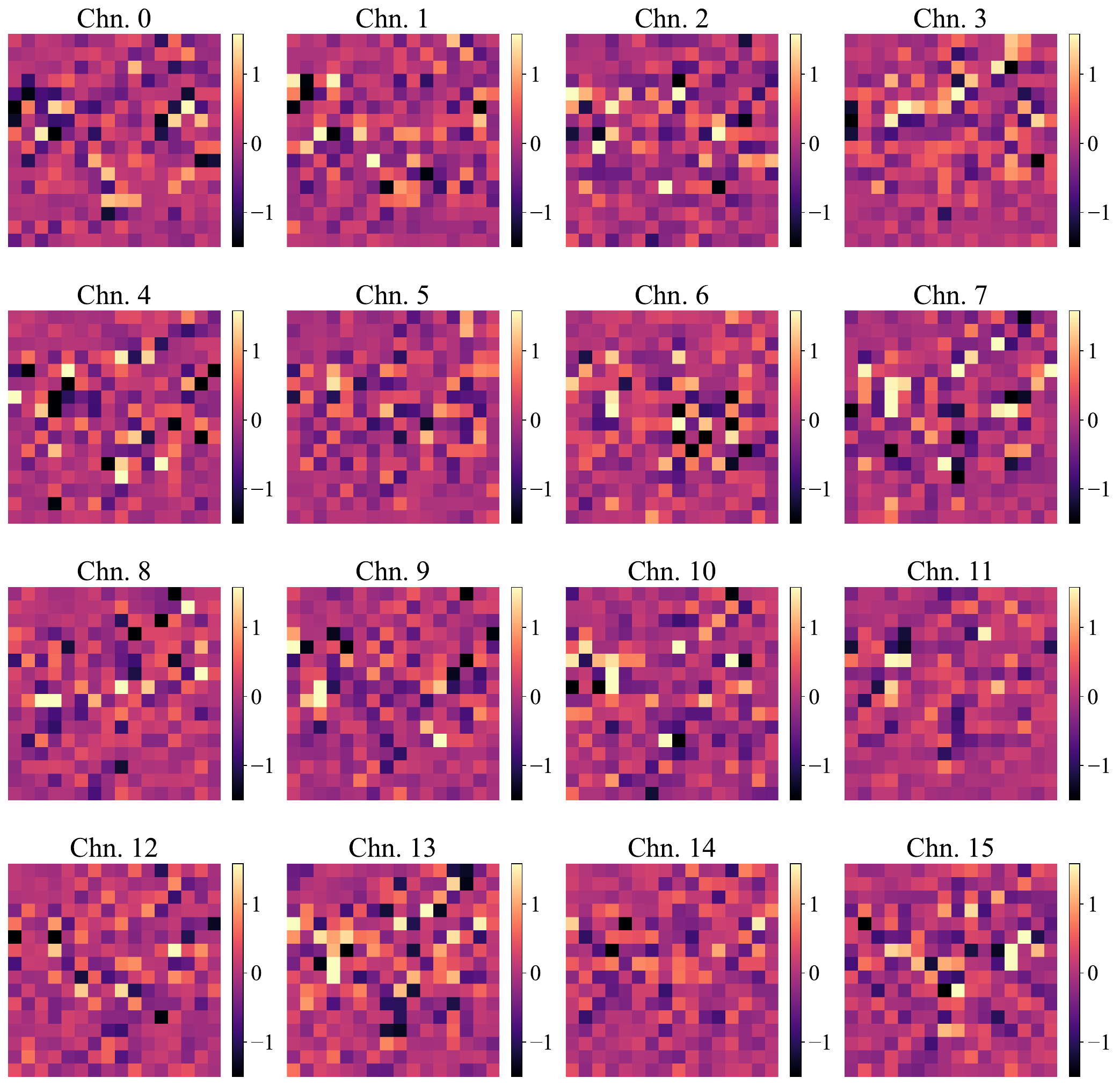}
        \caption{LHH}
    \end{subfigure}
    \caption{\textbf{Visualization of low-frequency wavelet subbands.}
    Low-frequency components exhibit smooth spatial variations and clear
    structural patterns, encoding the majority of semantic content.
    Diverse per-channel activation patterns suggest that each channel captures
    distinct semantic factors.}
    \label{fig:wavelet_low_freq}
\end{figure*}

\begin{figure*}[t]
    \centering
    \begin{subfigure}[b]{0.49\textwidth}
        \centering
        \includegraphics[width=\linewidth]{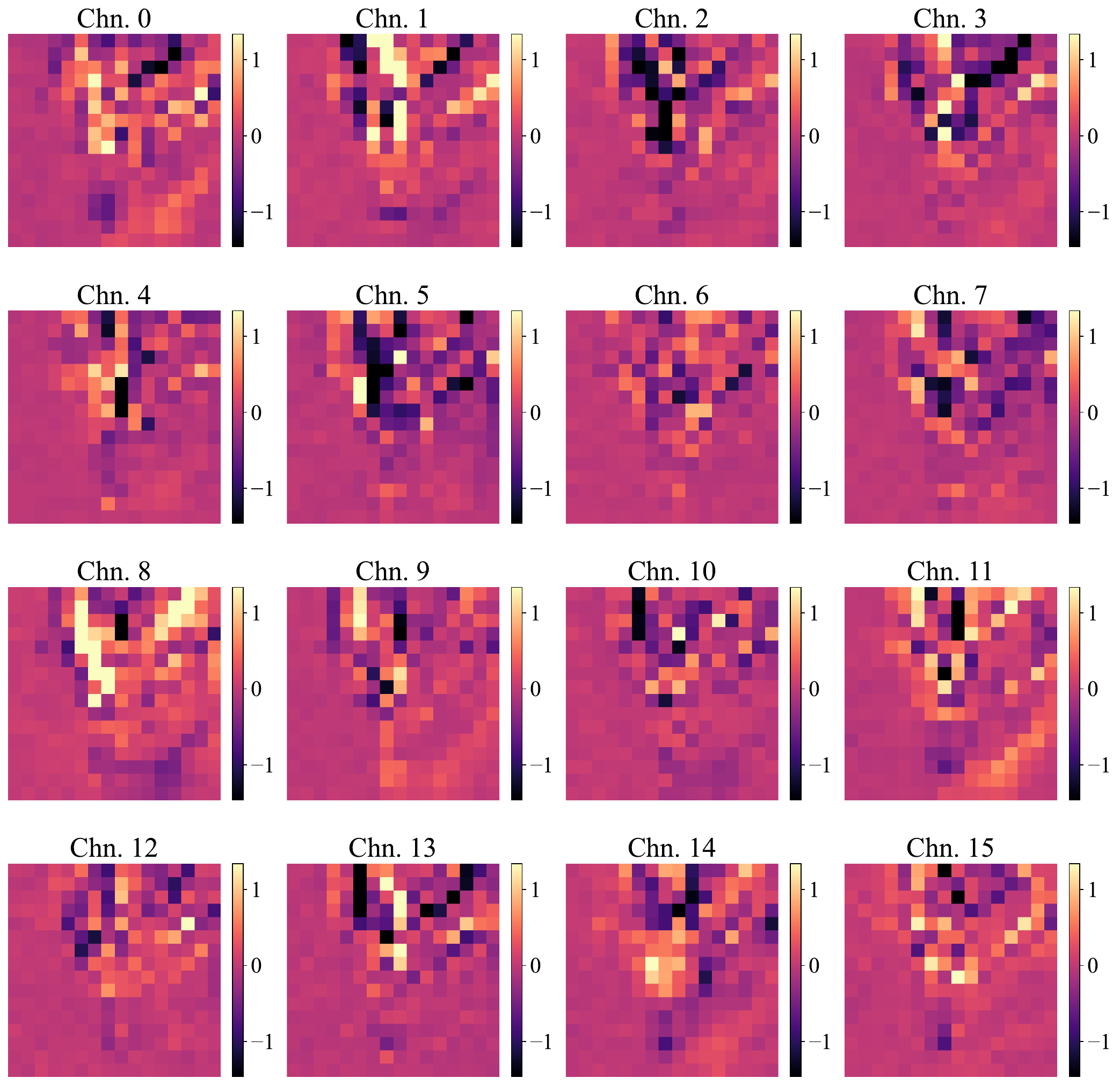}
        \caption{HLL}
    \end{subfigure}
    \hfill
    \begin{subfigure}[b]{0.49\textwidth}
        \centering
        \includegraphics[width=\linewidth]{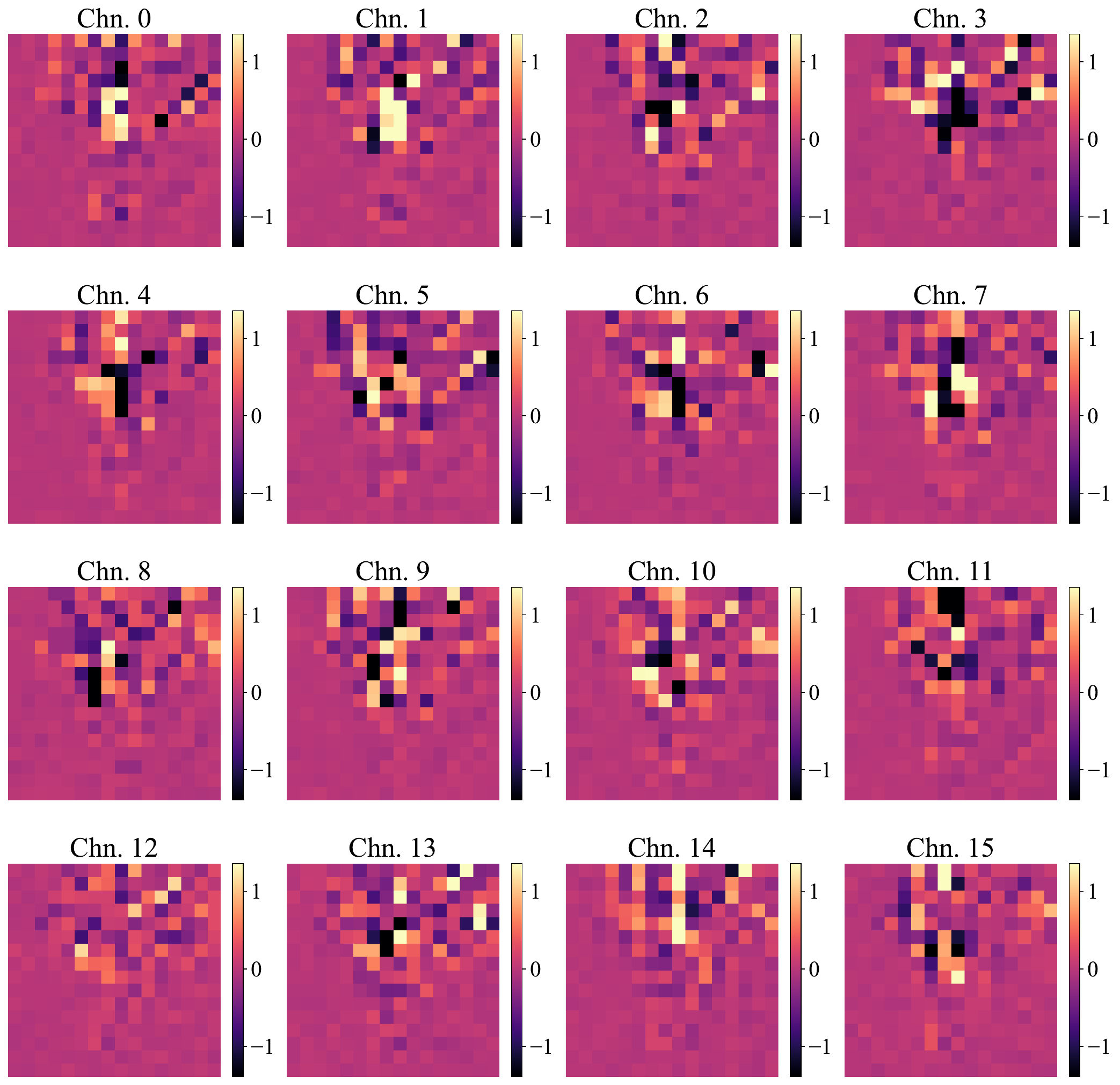}
        \caption{HLH}
    \end{subfigure}
    \vspace{0.6em}
    \begin{subfigure}[b]{0.49\textwidth}
        \centering
        \includegraphics[width=\linewidth]{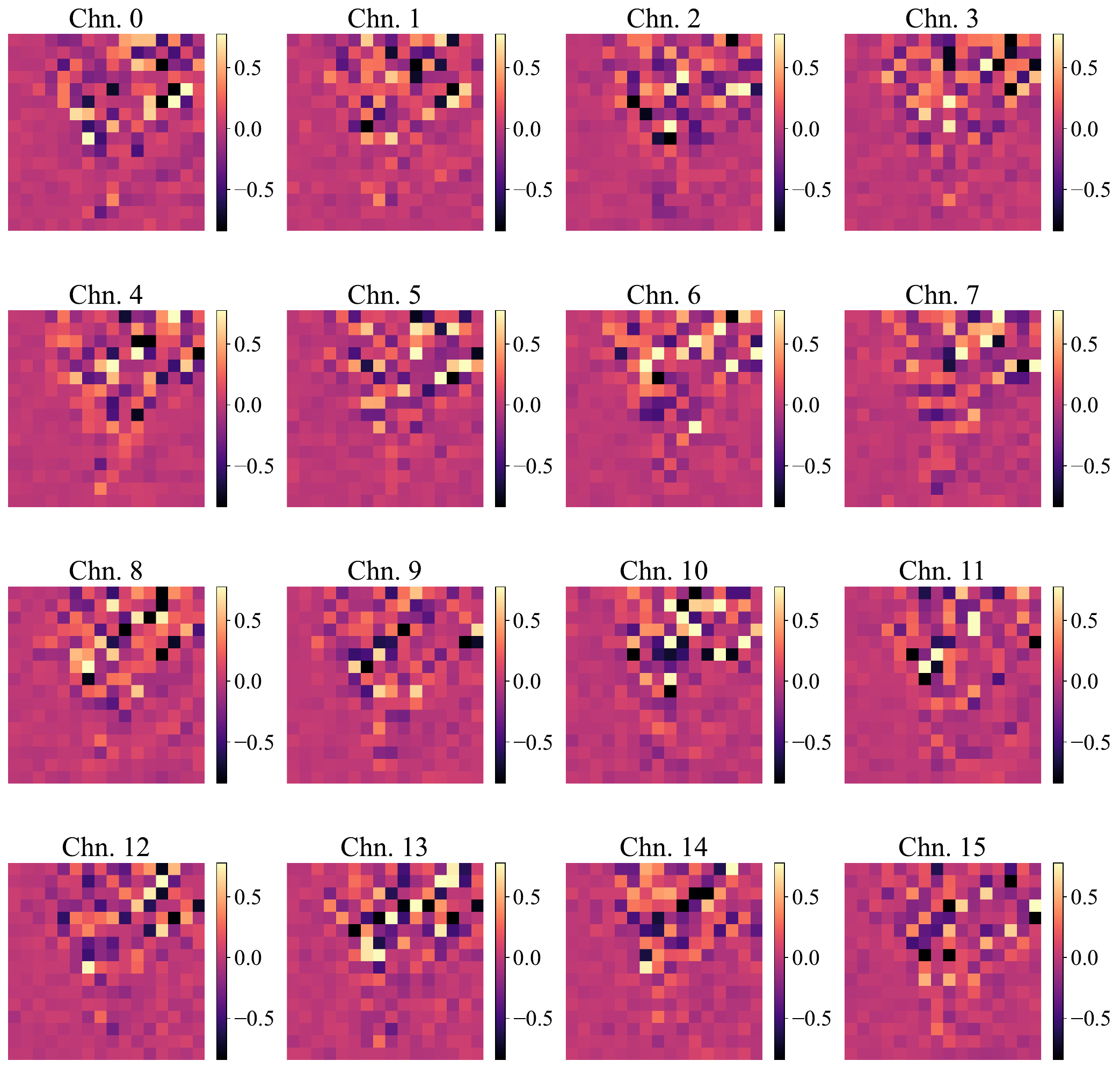}
        \caption{HHL}
    \end{subfigure}
    \hfill
    \begin{subfigure}[b]{0.49\textwidth}
        \centering
        \includegraphics[width=\linewidth]{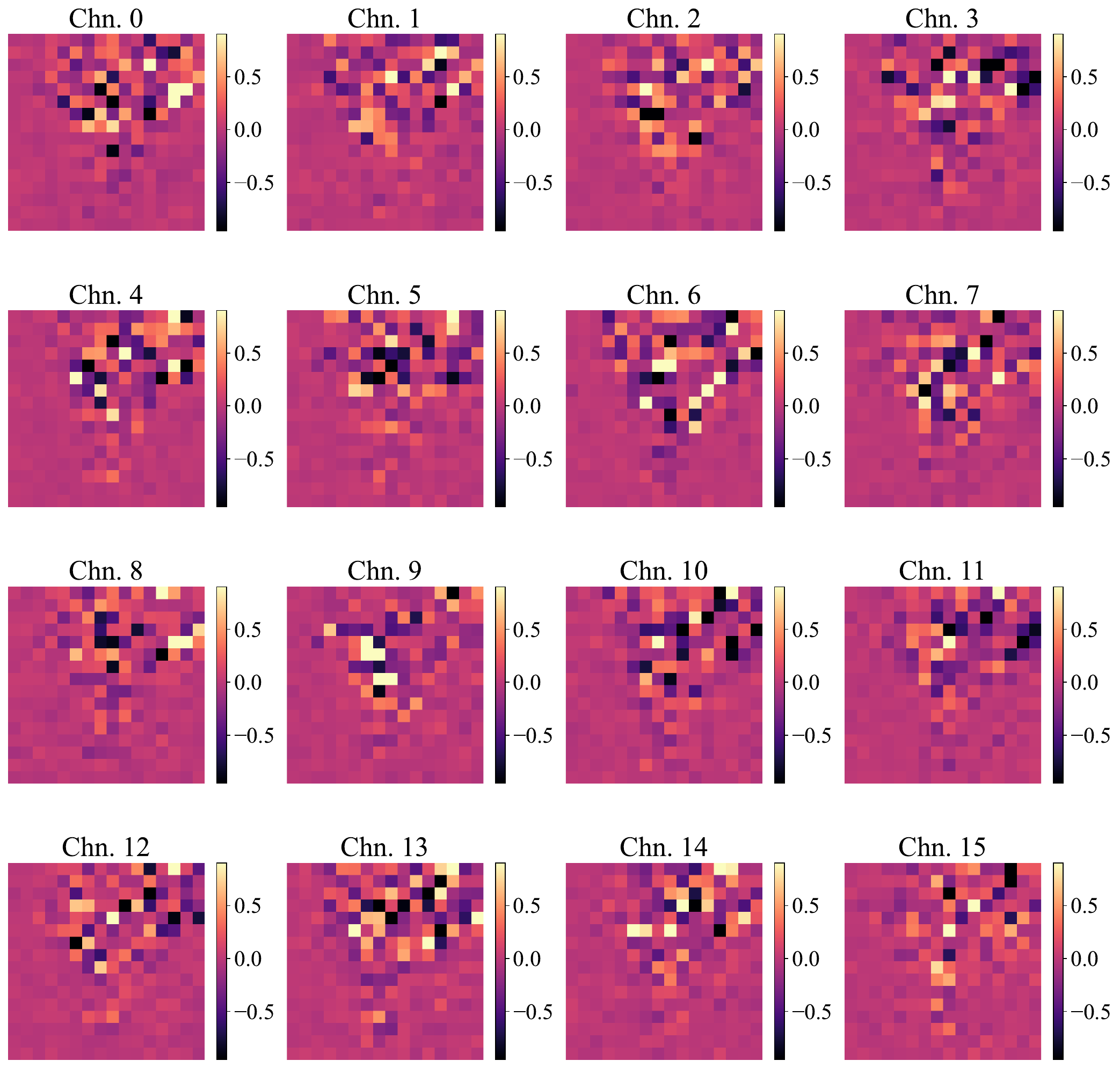}
        \caption{HHH}
    \end{subfigure}
    \caption{\textbf{Visualization of high-frequency wavelet subbands.}
    High-frequency components contain rapid local fluctuations with little
    channel-wise variation, resembling noise-like textures and contributing
    minimal semantic information.}
    \label{fig:wavelet_high_freq}
\end{figure*}

Given a latent tensor of shape $(T\!\times\!C\!\times\!H\!\times\!W)$ produced
by WF-VAE~\cite{li2025wf}, we apply a 3D WT along the temporal and spatial
dimensions to obtain eight subbands.
As shown in \cref{fig:wavelet_low_freq,fig:wavelet_high_freq}, the
low-frequency subbands contain rich, semantically diverse content, whereas
high-frequency subbands are nearly identical across channels—indicating that
high-frequency components are amenable to further compression.

\subsection{Subband Energy Distribution}
\label{sec:energy}

\begin{figure}[ht]
    \centering\footnotesize
    \includegraphics[width=\linewidth,
        trim={0 50pt 0 50pt},clip]{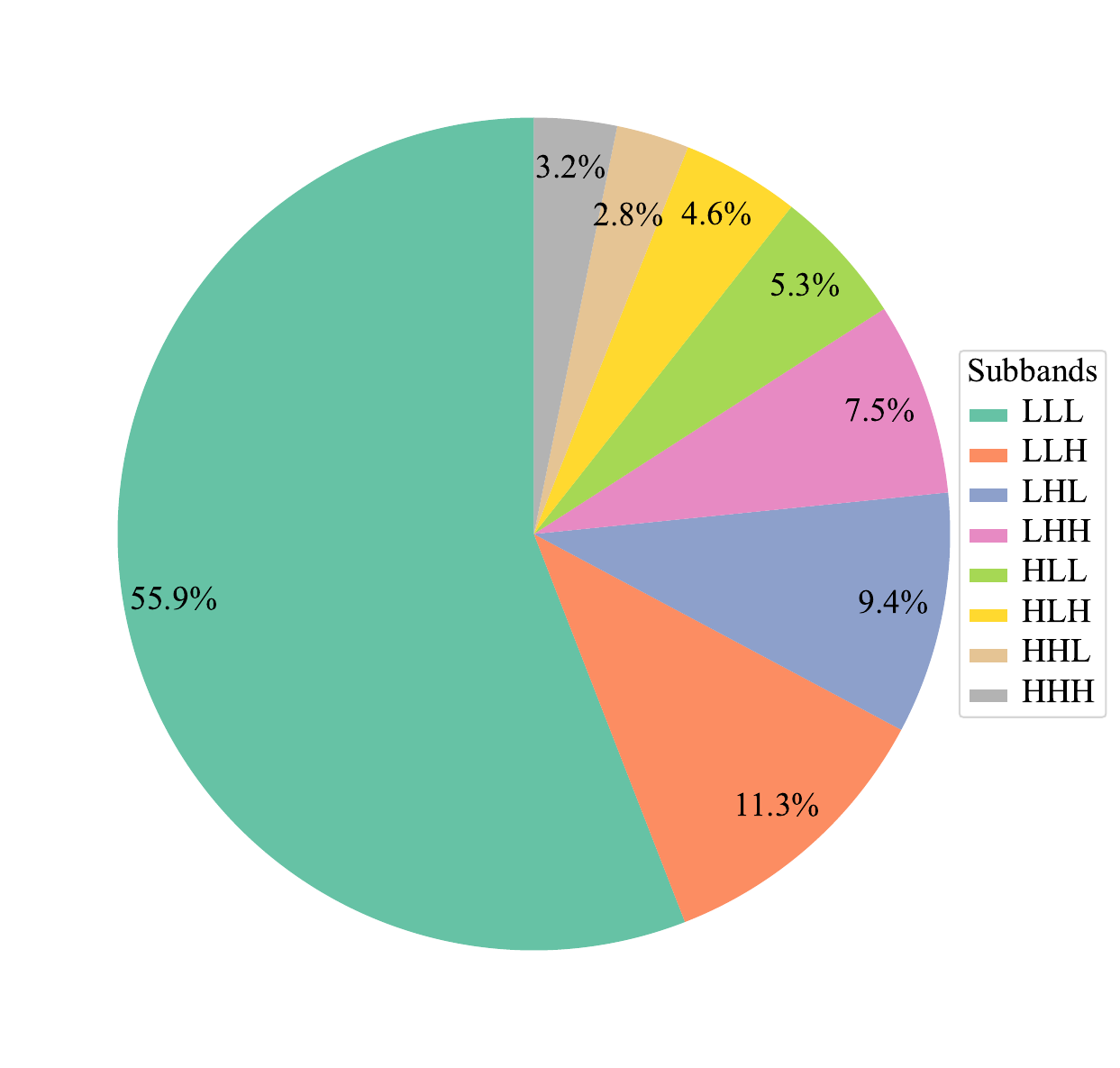}
    \caption{\textbf{Overall energy distribution across wavelet subbands
    (WebVid-10M).}
    Low-frequency subbands dominate, accounting for $\sim$85\% of total energy.}
    \label{fig:energy_pie}
\end{figure}

\begin{figure}[ht]
    \centering\footnotesize
    \includegraphics[width=\linewidth,
        trim={0 0 0 0},clip]{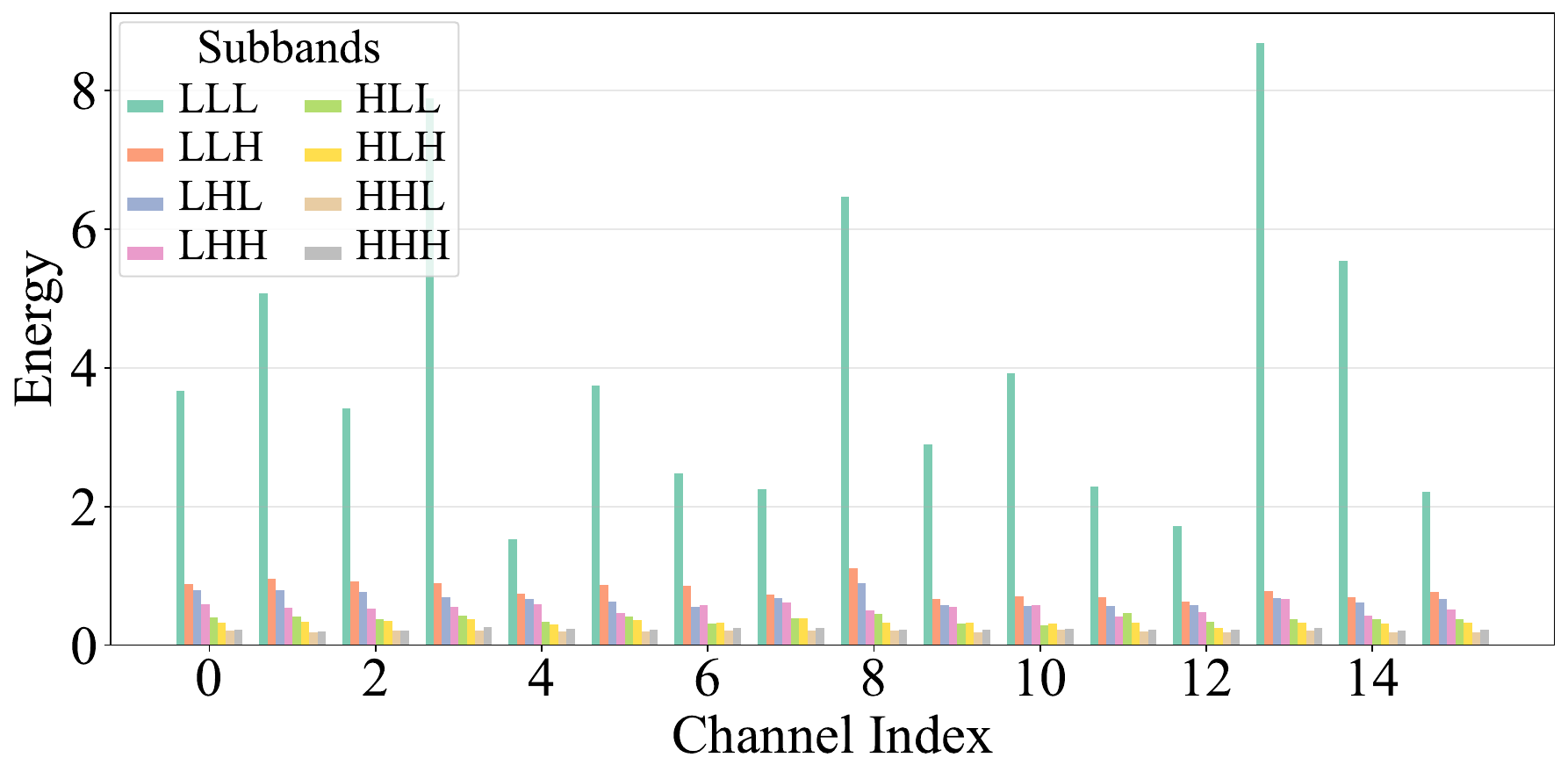}
    \caption{\textbf{Per-channel energy distribution across wavelet subbands.}
    The LLL subband shows substantially larger channel-wise variation than all
    high-frequency subbands, which exhibit nearly uniform energy across channels.}
    \label{fig:enegy_bar}
\end{figure}

We compute the average energy (mean squared magnitude) of each subband over
WebVid-10M. Results in \cref{fig:energy_pie,fig:enegy_bar} show that the
low-frequency subbands (LLL, LLH, LHL, LHH) collectively account for roughly
85\% of the total latent energy, while the LLL subband additionally exhibits
the greatest channel-wise variation. The remaining subbands show nearly uniform
per-channel energy, confirming that they contribute little semantic information.

\subsection{Lag-1 Temporal Autocorrelation}
\label{sec:autocorr}

To quantify temporal smoothness, we compute the lag-1 temporal
autocorrelation~\cite{sun2025adaptingllmstimeseries} for each subband.
Given a temporal sequence $\{x_t\}_{t=1}^{T}$:
\begin{equation}
\rho(1)
= \frac{\mathbb{E}[(x_t-\mu)(x_{t+1}-\mu)]}{\sigma^{2}},
\end{equation}
where $\mu=\mathbb{E}[x_t]$ and $\sigma^{2}=\mathbb{E}[(x_t-\mu)^2]$.
For multi-dimensional tensors of shape $T\!\times\!C\!\times\!H\!\times\!W$:
\begin{equation}
\rho_c(1)
= \frac{\mathbb{E}_{t,h,w}[(x_{t,c,h,w}-\mu_c)(x_{t+1,c,h,w}-\mu_c)]}
       {\sigma_c^{2}},
\end{equation}
where $c$ indexes channels.
Results (visualised in \cref{fig:frequency_analysis}) show that
low-frequency subbands (LLL, LLH) consistently exhibit higher autocorrelation
than high-frequency subbands (HHL, HHH), confirming that low-frequency
components encode temporally stable structures whereas high-frequency
components capture rapidly varying, noise-like details.

This observation also explains the improved zero-shot generalisation of LC-VAE
(see \cref{tab:zero_shot}): low-frequency content such as global layout and
coarse motion is consistent across datasets, whereas high-frequency details are
domain-specific and difficult to transfer. By discarding these high-frequency
variations, LC-VAE concentrates the latent space on the most stable and
semantically relevant components.

\section{Additional Ablation Studies}
\label{sec:ablation}

\subsection{Adaptive vs.\ Fixed Frequency Selection}
\label{sec:adaptive_ablation}

\begin{figure}[ht]
  \centering
  \includegraphics[width=0.96\linewidth]{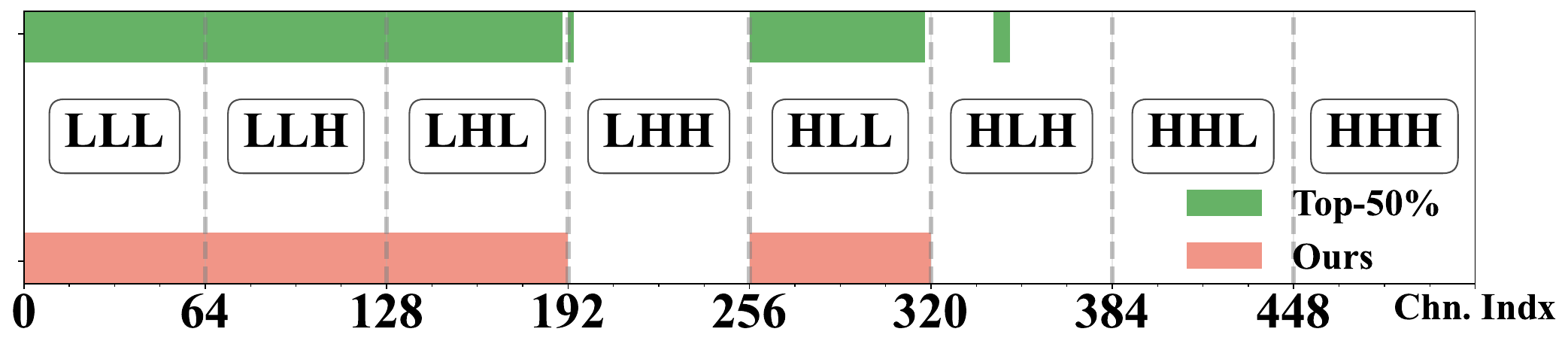}
  \caption{\textbf{Selected channels under adaptive Top-50\% selection.}
  After 20k training steps the scheme converges to 98\% channel overlap with
  our fixed design.}
  \label{fig:channel_distribution}
\end{figure}

A natural alternative to our fixed zero-out design is to learn which channels
to retain adaptively. We implement an \emph{adaptive Top-50\%} baseline that,
at each training step, preserves the 50\% of channels with the highest energy.
As shown in \cref{fig:channel_distribution}, after 20k steps this scheme
converges to the same subbands (LLL, LLH, LHL, HLL) as our fixed design, with
a 98\% channel overlap—empirically validating that our fixed mask closely
approximates the data-driven optimum.

Despite this convergence, the adaptive baseline performs worse in reconstruction
(\cref{tab:adaptive_vs_fixed}): learning the energy distribution itself
introduces additional training overhead that reduces effective capacity under a
fixed compute budget. Our fixed design avoids this cost, consistent with
classical approaches such as JPEG that zero out high-frequency coefficients
directly via quantization.

\begin{table}[ht]
\centering
\caption{\textbf{Reconstruction: adaptive Top-50\% vs.\ fixed design.}}
\label{tab:adaptive_vs_fixed}
\footnotesize
\setlength{\tabcolsep}{3.5pt}
\renewcommand{\arraystretch}{1.12}
\resizebox{\columnwidth}{!}{
\begin{tabular}{l c cccc cccc}
\toprule
\multirow{2}{*}{\textbf{Method}} & \multirow{2}{*}{\textbf{Chn.}}
& \multicolumn{4}{c}{\textbf{WebVid-10M}}
& \multicolumn{4}{c}{\textbf{OpenVid-1M}} \\
\cmidrule(lr){3-6}\cmidrule(lr){7-10}
& & \textbf{PSNR}$\uparrow$ & \textbf{SSIM}$\uparrow$ & \textbf{LPIPS}$\downarrow$ & \textbf{rFVD}$\downarrow$
  & \textbf{PSNR}$\uparrow$ & \textbf{SSIM}$\uparrow$ & \textbf{LPIPS}$\downarrow$ & \textbf{rFVD}$\downarrow$ \\
\midrule
Top-50\%       & 8  & 30.37 & 0.8456 & 0.0536 & 379.46
               &       32.13 & 0.8329 & 0.0495 & 257.81 \\
LC-VAE (Ours)  & 8  & \textbf{30.51} & \textbf{0.8637} & \textbf{0.0485} & \textbf{352.73}
               &       \textbf{32.24} & \textbf{0.8746} & \textbf{0.0414} & \textbf{234.10} \\
\midrule
Top-50\%       & 16 & 31.49 & 0.8597 & 0.0425 & 337.85
               &       33.18 & 0.8354 & 0.0384 & 181.61 \\
LC-VAE (Ours)  & 16 & \textbf{31.81} & \textbf{0.8839} & \textbf{0.0413} & \textbf{319.24}
               &       \textbf{33.26} & \textbf{0.8973} & \textbf{0.0329} & \textbf{168.46} \\
\bottomrule
\end{tabular}}
\end{table}

\subsection{Robustness on Challenging Video Content}
\label{sec:challenging}

We construct \emph{texture-heavy} and \emph{fast-motion} subsets from
OpenVid-1M and UCF-101 to stress-test the method. The texture-heavy subset
contains clips with rich fine-grained details (e.g., foliage, fabrics); the
fast-motion subset contains clips with large inter-frame displacements.
As shown in \cref{tab:challenging}, LC-VAE consistently outperforms WF-VAE
across all metrics on both subsets, demonstrating robustness to challenging
spatial textures and rapid temporal dynamics.

\begin{table}[ht]
\centering
\caption{\textbf{Reconstruction on texture-heavy and fast-motion subsets
(16-channel VAEs).}}
\label{tab:challenging}
\footnotesize
\setlength{\tabcolsep}{3.5pt}
\renewcommand{\arraystretch}{1.12}
\resizebox{\columnwidth}{!}{
\begin{tabular}{l c cccc cccc}
\toprule
\multirow{2}{*}{\textbf{Method}} & \multirow{2}{*}{\textbf{Chn.}}
& \multicolumn{4}{c}{\textbf{Texture-Heavy}}
& \multicolumn{4}{c}{\textbf{Fast-Motion}} \\
\cmidrule(lr){3-6}\cmidrule(lr){7-10}
& & \textbf{PSNR}$\uparrow$ & \textbf{SSIM}$\uparrow$ & \textbf{LPIPS}$\downarrow$ & \textbf{rFVD}$\downarrow$
  & \textbf{PSNR}$\uparrow$ & \textbf{SSIM}$\uparrow$ & \textbf{LPIPS}$\downarrow$ & \textbf{rFVD}$\downarrow$ \\
\midrule
WF-VAE        & 16 & 36.14 & 0.9587 & 0.0153 & 124.95
              &       33.64 & 0.9431 & 0.0206 & 195.78 \\
LC-VAE (Ours) & 16 & \textbf{36.78} & \textbf{0.9622} & \textbf{0.0142} & \textbf{112.78}
              &       \textbf{34.07} & \textbf{0.9435} & \textbf{0.0194} & \textbf{177.62} \\
\bottomrule
\end{tabular}}
\end{table}

\subsection{Effect of Data Quality on rFVD}
\label{sec:rfvd_analysis}

In the main paper, rFVD for LC-VAE occasionally regresses on lower-quality
datasets (WebVid-10M, UCF-101) despite consistent PSNR gains. We hypothesize
that LC-VAE acts as an implicit artifact filter: suppressing high-frequency
latent components discourages the model from fitting compression noise present
in degraded data.

To verify this, we construct \emph{OpenVid-1M (Compressed)} by re-encoding the
original clips with H.264 to simulate real-world transmission quality loss.
We then evaluate reconstruction against both the compressed inputs and the
original clean reference frames. As shown in \cref{tab:rfvd_quality}, LC-VAE
yields \emph{worse} rFVD relative to the compressed inputs (it does not overfit
to artifacts) but \emph{better} rFVD relative to the clean originals,
confirming that our method implicitly filters artifacts while preserving
perceptual fidelity on clean data.

\begin{table}[ht]
\centering
\caption{\textbf{rFVD evaluated against compressed vs.\ clean references
(16-channel, OpenVid-1M).}}
\label{tab:rfvd_quality}
\footnotesize
\setlength{\tabcolsep}{3.5pt}
\renewcommand{\arraystretch}{1.12}
\resizebox{\columnwidth}{!}{
\begin{tabular}{l c cccc cccc}
\toprule
\multirow{2}{*}{\textbf{Method}} & \multirow{2}{*}{\textbf{Chn.}}
& \multicolumn{4}{c}{\textbf{OpenVid-1M (Compressed)}}
& \multicolumn{4}{c}{\textbf{OpenVid-1M (Original)}} \\
\cmidrule(lr){3-6}\cmidrule(lr){7-10}
& & \textbf{PSNR}$\uparrow$ & \textbf{SSIM}$\uparrow$ & \textbf{LPIPS}$\downarrow$ & \textbf{rFVD}$\downarrow$
  & \textbf{PSNR}$\uparrow$ & \textbf{SSIM}$\uparrow$ & \textbf{LPIPS}$\downarrow$ & \textbf{rFVD}$\downarrow$ \\
\midrule
WF-VAE        & 16 & 35.97 & 0.9297 & 0.0176 & \textbf{47.91}
              &       35.63 & 0.9308 & 0.0181 & 53.33 \\
LC-VAE (Ours) & 16 & \textbf{36.95} & \textbf{0.9375} & \textbf{0.0163} & 58.79
              &       \textbf{36.92} & \textbf{0.9384} & \textbf{0.0177} & \textbf{40.34} \\
\bottomrule
\end{tabular}}
\end{table}

\section{Extended Experiments}
\label{sec:extended}

\subsection{Generation with Longer Training}
\label{sec:longer_training}

The generation experiments in the main paper use 100k diffusion training steps.
Here we extend to 300k steps with Latte-L~\cite{ma2025latte} trained on
OpenVid-1M and WebVid-10M (4$\times$ NVIDIA H200 GPUs, $\approx$4 days).
As shown in \cref{tab:gen_300k}, LC-VAE consistently outperforms WF-VAE across
all channel counts, and the advantage grows with more channels, confirming that
larger-channel LC-VAEs benefit from additional diffusion training.

\begin{table}[ht]
\centering
\caption{\textbf{Video generation FVD$_{16}$ ($\downarrow$) with Latte-L,
300k training steps.}}
\label{tab:gen_300k}
\setlength{\tabcolsep}{10pt}
\renewcommand{\arraystretch}{1.15}
\resizebox{\columnwidth}{!}{
\begin{tabular}{l c c c}
\toprule
\textbf{VAE Method} & \textbf{Chn.}
& \textbf{WebVid-10M} & \textbf{OpenVid-1M} \\
\midrule
WF-VAE        & 4  & 492.87 & 279.65 \\
LC-VAE (Ours) & 4  & \textbf{473.46} & \textbf{229.03} \\
\midrule
WF-VAE        & 8  & 465.91 & 268.28 \\
LC-VAE (Ours) & 8  & \textbf{431.44} & \textbf{226.45} \\
\midrule
WF-VAE        & 16 & 457.22 & 236.42 \\
LC-VAE (Ours) & 16 & \textbf{414.38} & \textbf{217.51} \\
\bottomrule
\end{tabular}}
\end{table}

\subsection{Scalability to Larger Architectures}
\label{sec:scalability}

\paragraph{WanVAE-2.1.}
We apply our latent compression to WanVAE-2.1~\cite{wan2025wan} to verify
generality beyond WF-VAE. As shown in \cref{tab:wan_vae_recon}, LC-VAE
consistently improves PSNR and rFVD over the vanilla baseline on both datasets,
demonstrating that fixed high-frequency zero-out is architecture-agnostic.

\begin{table}[ht]
\centering
\caption{\textbf{Reconstruction with LC-VAE applied to WanVAE-2.1 (16 channels).}}
\label{tab:wan_vae_recon}
\footnotesize
\setlength{\tabcolsep}{3.5pt}
\renewcommand{\arraystretch}{1.12}
\resizebox{\columnwidth}{!}{
\begin{tabular}{l c cccc cccc}
\toprule
\multirow{2}{*}{\textbf{Method}} & \multirow{2}{*}{\textbf{Chn.}}
& \multicolumn{4}{c}{\textbf{WebVid-10M}}
& \multicolumn{4}{c}{\textbf{OpenVid-1M}} \\
\cmidrule(lr){3-6}\cmidrule(lr){7-10}
& & \textbf{PSNR}$\uparrow$ & \textbf{SSIM}$\uparrow$ & \textbf{LPIPS}$\downarrow$ & \textbf{rFVD}$\downarrow$
  & \textbf{PSNR}$\uparrow$ & \textbf{SSIM}$\uparrow$ & \textbf{LPIPS}$\downarrow$ & \textbf{rFVD}$\downarrow$ \\
\midrule
WanVAE-2.1 (Vanilla) & 16 & 29.26 & 0.8376 & \textbf{0.0621} & 381.19
                     &       30.99 & 0.8508 & 0.0551 & 274.05 \\
WanVAE-2.1 (Ours)    & 16 & \textbf{29.91} & \textbf{0.8437} & 0.0641 & \textbf{342.68}
                     &       \textbf{31.28} & \textbf{0.8591} & \textbf{0.0532} & \textbf{249.92} \\
\bottomrule
\end{tabular}}
\end{table}

\paragraph{Wan-2.1 (2.1B) Diffusion Model.}
We further train the Wan-2.1 diffusion model scaled to 2.1B parameters for
160k steps on WebVid-10M (8$\times$ H200 GPUs, $\approx$3 days).
\cref{tab:wan_gen} shows that LC-VAE substantially outperforms WF-VAE,
confirming that the benefits of latent compression persist at larger model
scales.

\begin{table}[ht]
\centering
\caption{\textbf{Video generation FVD$_{16}$ ($\downarrow$) with Wan-2.1 (2.1B)
on WebVid-10M.}}
\label{tab:wan_gen}
\setlength{\tabcolsep}{20pt}
\renewcommand{\arraystretch}{1.15}
\resizebox{\columnwidth}{!}{
\begin{tabular}{l c c}
\toprule
\textbf{Method} & \textbf{Chn.} & \textbf{WebVid-10M} \\
\midrule
WF-VAE        & 16 & 487.15 \\
LC-VAE (Ours) & 16 & \textbf{434.41} \\
\bottomrule
\end{tabular}}
\end{table}

\section{Additional Qualitative Results}
\label{sec:qualitative}

\begin{figure*}[t]
    \centering\footnotesize
    \includegraphics[width=\linewidth, trim={0 49mm 0 0mm}, clip]{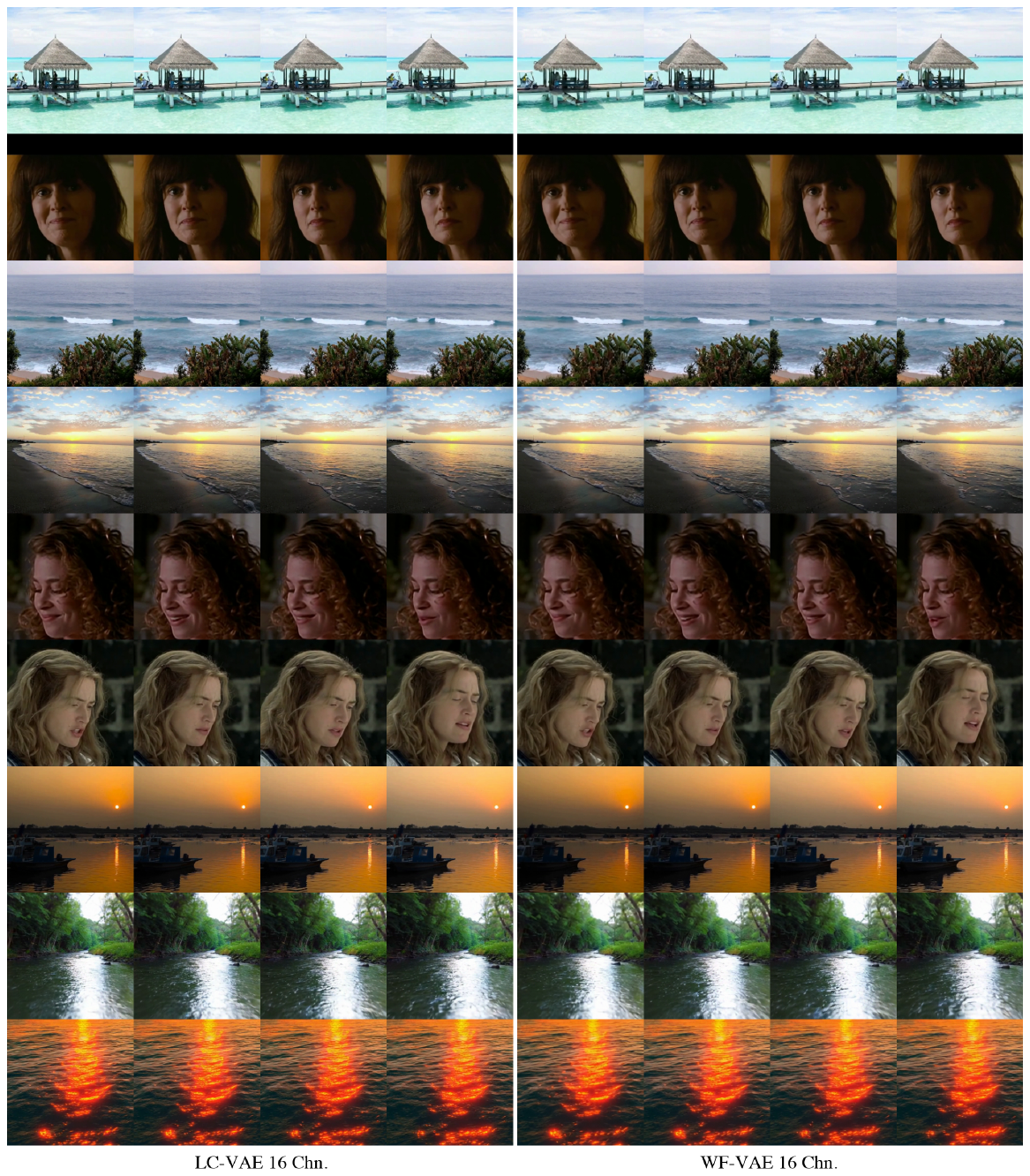}
    \vspace{-0.2em}
    \makebox[\linewidth][c]{%
      \begin{minipage}{0.35\linewidth}\centering LC-VAE 16 Chn.\end{minipage}%
      \hspace{0.11\linewidth}%
      \begin{minipage}{0.35\linewidth}\centering WF-VAE 16 Chn.\end{minipage}%
    }
    \vspace{0.2em}
    \caption{\textbf{Non-curated reconstruction on OpenVid-1M.}
    LC-VAE (left) vs.\ WF-VAE (right).}
    \label{fig:reconstruction}
\end{figure*}

\paragraph{Non-curated reconstruction.}
\cref{fig:reconstruction} shows non-curated reconstruction results produced by
LC-VAE on OpenVid-1M.

\begin{figure*}[t]
    \centering\footnotesize
    \includegraphics[width=\linewidth, trim={0 49mm 0 0mm}, clip]{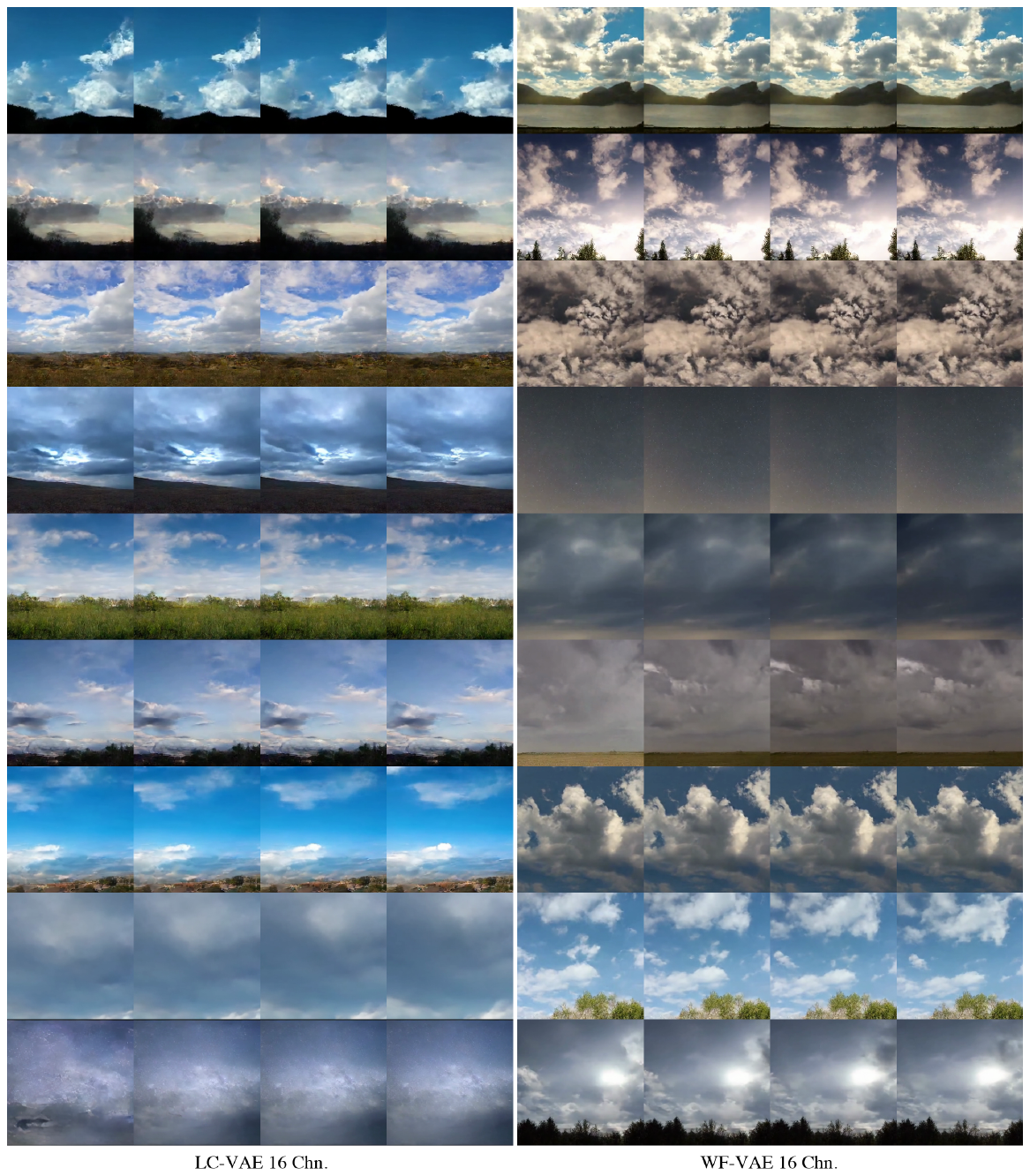}
    \vspace{-0.2em}
    \makebox[\linewidth][c]{%
      \begin{minipage}{0.35\linewidth}\centering LC-VAE 16 Chn.\end{minipage}%
      \hspace{0.11\linewidth}%
      \begin{minipage}{0.35\linewidth}\centering WF-VAE 16 Chn.\end{minipage}%
    }
    \vspace{0.2em}
    \caption{\textbf{Non-curated video generation on SkyTimelapse.}
    Latte~\cite{ma2025latte} under guidance-free sampling trained with
    LC-VAE (left) vs.\ WF-VAE (right).}
    \label{fig:generation}
\end{figure*}

\paragraph{Non-curated video generation.}
\cref{fig:generation} shows non-curated generation results obtained by
integrating LC-VAE into a latent diffusion video generator, evaluated on the
SkyTimelapse dataset.


\end{document}